\documentclass[11pt]{article}

\usepackage[preprint]{acl}

\usepackage{times}
\usepackage{latexsym}
\usepackage{amsmath}
\usepackage{xcolor}
\usepackage{tabularx}
\usepackage[T1]{fontenc}

\usepackage[utf8]{inputenc}

\usepackage{microtype}

\usepackage{inconsolata}

\usepackage{graphicx}

\usepackage{ulem}
\usepackage{microtype}
\usepackage{hyperref}
\usepackage{url}
\usepackage{booktabs}
\usepackage{arydshln}
\usepackage{wrapfig}
\usepackage{algorithm}
\usepackage{algpseudocode}
\usepackage{graphicx}
\usepackage{enumitem}
\title{Inter-dimension Dependence for Multi-Dimensional Evaluation of Open-Ended Text}

\author{
  \textbf{Haoyuan Li}, 
  \textbf{Snigdha Chaturvedi}
\\
  University of North Carolina at Chapel Hill
\\
  {\{haoyuanl, snigdha\}@cs.unc.edu}
}

\begin{document}
\maketitle
\begin{abstract}
LLM-as-a-judge methods are widely used for evaluating the quality of generated open-ended text. Such evaluations are generally multi-dimensional, since the error patterns in texts can be different for different dimensions. Therefore, reliable LLM judges should evaluate each target dimension independently. To quantify the extent to which LLM judges depend on non-target dimensions when evaluating a target dimension, i.e., inter-dimension dependence, we propose CorrGap. To measure this, CorrGap uses the difference in correlations between LLM-predicted scores and ground truth scores across different groups of texts. Using CorrGap, we show that inter-dimension dependence is pervasive across LLM judges in open-ended text evaluation tasks. To mitigate inter-dimension dependence, we propose DimCheck, a method that iteratively removes unrelated evidence from COTs generated by LLM judges in a step-wise way. We show that DimCheck mitigates inter-dimension dependence and outperforms strong baselines across three LLMs and four tasks. We also show that smaller trained LLMs can approximate larger LLMs in DimCheck, with much lower inference costs.
\end{abstract}

\section{Introduction}

LLM-as-a-judge methods \citep{liu2023g, chiang2023closer, kim2024debate} are widely used for evaluating generated open-ended text. These methods assign either a single overall score to the text \citep{zheng2023judging, lambert2025rewardbench} or multiple scores along multiple dimensions. In the latter setting, referred to as multi-dimensional evaluation, each dimension assesses a specific aspect of the text and the text is generally evaluated along one dimension at a time. 
For example, summaries are often evaluated across dimensions like relevance to the input document(s), factuality and fluency \citep{fabbri2020summeval}. Similarly, automatically generated narratives are evaluated on relevance to the prompt, coherence, etc. This multi-dimensional evaluation is preferable because open-ended texts can show different error patterns for different dimensions 
\citep{hoskinghuman,bao2026position}. Aggregating these errors into a single overall score may overlook important deficiencies since errors in different dimensions are not directly comparable or interchangeable. For example, a fluent summary with factual errors and a factually correct summary with fluency errors can receive the same overall score, but reflect fundamentally different error patterns. 
Multi-dimensional evaluation can effectively capture these distinct error patterns and offer greater interpretability. In these settings, it is important for LLM judges to evaluate each target dimension independently, the evaluation may effectively collapse into a single overall score, reducing 
the intended benefits of multi-dimensional rewards.

\begin{table}
\scriptsize
\centering
\begin{tabular}{|p{6.7cm}|}
\hline
\textbf{Target Dimension}: \textbf{Relevance} \\
\textbf{Non-target Dimension}: \textbf{Consistency, Coherence, Fluency} \\ 
The summary includes some important information such as Ze Maria's representation of Brazil and his role as the third coach at Ceahlaul. However, it is missing key details about the timeline of events (e.g., being reinstated and then fired again) and the context of the club's relegation threat. \textcolor{red}{Additionally, the summary is somewhat disjointed and lacks a clear narrative flow.} It also omits the reason for his second firing and the replacement by Vanya Radinovic. These omissions reduce the relevance. \\
\hline
\end{tabular}
\caption{An example COT generated by Llama3.3-70B-Instruct as a judge when evaluating a summary on relevance (the target dimension). The text in \textcolor{red}{red} is related to a non-target dimension, coherence, but not relevance.}
\label{tab:example_cot}
\end{table}
However, we hypothesize that LLM judges cannot always evaluate along different dimensions independently. Tab. \ref{tab:example_cot} shows the Chain-of-Thought (COT) reasoning of an LLM judge when evaluating the relevance of a summary (\textit{target dimension}). It incorrectly depends on evidence related to coherence, a \textit{non-target dimension}(highlighted in red). To quantify the extent to which LLM judges depend on non-target dimensions when evaluating the text on a target dimension, i.e., \textit{inter-dimension dependence}, we propose CorrGap. CorrGap is a novel evaluation metric based on the assumption that, if the LLM judge is not able to do independent evaluations, i.e. inter-dimension dependence exists, the judge's score for the target dimension will be highly influenced by the text's quality on other dimensions. This will be especially damaging on texts for which the ground truth score for the target dimension is very different from the score for other dimensions. In other words, the LLM judge will be especially incorrect on these texts. 
We evaluate CorrGap in two different setups using three different LLM judges and find it to be a reliable measure of inter-dimensional dependence.  Using CorrGap, we measure the inter-dimension dependence with three evaluation frameworks using eight LLMs for four open-ended text generation tasks. Across most settings, we observe significantly higher CorrGap scores indicating pervasive inter-dimension dependence in LLM judges. 


To mitigate inter-dimension dependence, we present DimCheck that analyzes the Chain-Of-Thought (COT) \citep{wei2022chain} generated by LLM judges before rating. It iteratively removes unrelated evidence (such as those highlighted in red in Tab.~\ref{tab:example_cot}) from the original COT and performs evaluation again conditioned on the modified COT.

We evaluate DimCheck by performing comprehensive experiments across three LLMs and four tasks. The results show that DimCheck outperforms other strong baselines. We further show that smaller LLMs can be trained to approximate larger LLMs in DimCheck with reduced inference cost. 

Our contributions are three-fold: 
\begin{itemize}[topsep=1pt, leftmargin=*, noitemsep]
\item CorrGap, a metric that quantitatively evaluates inter-dimension dependence for multi-dimensional open-ended text evaluation;
\item Comprehensive experiments that show inter-dimension dependence is significant across evaluation frameworks and tasks;
\item DimCheck, a method designed to mitigate inter-dimension dependence by removing unrelated evidence from the original COT. 
\end{itemize}

\section{Related Work}
Open-ended texts can show various error patterns \citep{hoskinghuman,bao2026position}. Therefore, they are generally evaluated along multiple dimensions \citep{fabbri2020summeval}. Beyond evaluation, multi-dimensional evaluation is also used to provide a comprehensive reward signal during alignment by mutli-objective alighment \citep{he2025pareto, yang2024rewards} or detailed rubrics \citep{gunjal2025rubrics,huang2025reinforcement}.

LLM judges \citep{wang2023chatgpt, fu2024gptscore} are widely used to evaluate the quality of texts in a pointwise \citep{gao2023human} or pairwise way \citep{zheng2023judging}. In this work, we focus on pointwise evaluation. To improve the correlation between predicted and ground truth scores, previous works 
model the uncertainty of predicted scores \citep{liu2023g}, analyze the text and instruction before evaluation \citep{chiang2023closer, wu-etal-2025-seeval, kim-etal-2024-prometheus}, enable interaction between agents \citep{chanchateval,kim2024debate,kumar2025courteval}, or incorporate a more detailed rubric \citep{kim2023prometheus,lee2025checkeval}. 
Although showing high correlation, such LLM judges can be susceptible to different types of biases \citep{ye2025justice} like self-enhancement bias \citep{goyal2022news} or verbosity bias \citep{zheng2023judging}. 
Existing studies typically measure such biases using perturbation-based methods. 
These biases can be viewed as non-target dimensions in our setting. However, perturbation-based evaluation requires carefully designed modifications for each dimension and can be difficult to construct for certain dimensions, such as relevance in summarization. 
Instead, \cite{xiao2023evaluating} show that LLM-predicted scores for a target dimension can correlate more with ground truth scores of non-target dimensions than with those of the target dimension for summarization. This hints that LLM judges might not be able to independently evaluate the target dimension. 
However, this correlation-based analysis remains coarse-grained and qualitative. Even if the predicted scores correlate more with the ground truth scores of the target dimension, such correlations alone do not guarantee that the evaluation of the target dimension is independent. 
\section{CorrGap} 
\label{sec:corrgap}
Multi-dimensional evaluation evaluate a set of texts $T$ along $n$ dimensions: $d_1,...,d_n$. For a text $t\in T$, an LLM judge assigns a score $\hat{s}_{i,t}$ and human annotators annotate a ground truth score $s_{i,t}$ on each dimension $d_i$. We define inter-dimension dependence as the extent to which the predicted score $\hat{s}_i$ on the target dimension $d_i$ is influenced by the non-target dimensions $\{d_{k\neq i}\}$

CorrGap is a metric designed for inter-dimension dependence motivated by stratified analysis \citep{cochran1954combination}. To understand the intuition of CorrGap, consider the following two groups of texts. The first group, the low-variance group, $T^i_l$, contains texts, $t$,  whose ground truth score $s_{i,t}$ on the target dimension, $d_i$,  are similar to the scores on other dimensions, $\{s_{k\neq i,t}\}$. The other group, the high-variance group, $T^i_h$, contains texts whose ground truth scores on the target dimension are different from non-target dimensions. If an LLM judge cannot evaluate dimensions independently, then its scores on the target dimension will be influenced by other dimensions, making the predicted scores less accurate. The accuracy will take a worse hit for the high than the low-variance group, because in the high-variance group, the scores for target dimensions should ideally be very different from the non-target dimensions (so getting influenced by non-target dimensions will be particularly problematic). CorrGap measures the LLM's inability to do independent evaluations by considering the difference in this accuracy on the low- and high- variance groups, $T^i_l$ and $T^i_h$. The accuracy is measured as the correlation of predicted scores with ground-truth scores.


However, distributions of ground truth scores across dimensions can be very different as in Fig. \ref{fig:distribution}. Therefore, CorrGap measures the differences between dimensions $\Delta(t,d_i)$ using the percentile $s^p_{i,t}$  of the score $s_{i,t}$ among all scores $\{s_{i,t'} ,t'\in T\}$ instead of its raw value. To construct the groups, for each text $t\in T$, CorrGap calculates the differences in the percentiles of ground truth scores on the target dimension  $d_i$ and other dimensions: $\Delta(t,d_i)=|s^p_{i,t}-mean(\{s^p_{k\neq i, t}\})|$.


Only using differences in percentiles of ground truth scores $\Delta(t,d_i)$ to split texts may lead to two groups with markedly different distributions of ground truth scores $s_{i,t}$ as in Tab. \ref{tab:kl_div}. This makes their correlations with LLM-predicted scores $\hat{s}_{i,t}$ incomparable. For example, one group can be concentrated around a certain value while the other spans a wider range. In this case, the differences in correlation can arise from the distribution difference rather than genuine inter-dimension dependence. Therefore, CorrGap splits texts within each possible value of the ground truth score $s_{i,t}$.\footnote{If ground truth scores are continuous, we can bin them.} 
Specifically, for a possible ground truth score value $k$, CorrGap considers the subset $\{t|s_{i,t}=k,t\in T\}$ and splits it by the differences in the percentiles $\Delta(t,d_i)$. Texts with the top $50$\% value of $\Delta(t,d_i)$ are assigned to the high-variance group $T^i_h$ and the remaining to the low-variance group $T^i_l$. 

For a dimension $d_i$, CorrGap is then defined as the difference in correlations between LLM-predicted scores $\hat{s}_i$ and ground truth scores $s_i$ across two groups $T^i_l$ and $T^i_h$:
\vspace{-0.05cm}
\begin{align}
CG_i=|corr(\{(s_{i,t},\hat{s}_{i,t})|t\in T^i_l \}) \notag \\
-corr(\{(s_{i,t},\hat{s}_{i,t})|t\in T^i_h \})|. 
\end{align}
If an LLM judge can independently evaluate texts on dimension $d_i$, the correlation on groups $T^i_l$ and $T^i_h$ should be similar, yielding a CorrGap value $CG_i$ close to zero. Conversely, if the judge cannot do independent evaluation, correlations in the high-variance group will be much lower than those in the low-variance group. Therefore, the CorrGap value $CG_i$ will be higher. We show that CorrGap is an accurate metric for inter-dimension dependence in Sec .\ref{sec:corrgap_effect} and show that inter-dimension dependence is pervasive for LLM judges in Sec. \ref{sec:corrgap_eval}. 

\section{DimCheck}
\begin{figure*}[t]
\centering
\includegraphics[width=0.8\textwidth,keepaspectratio]{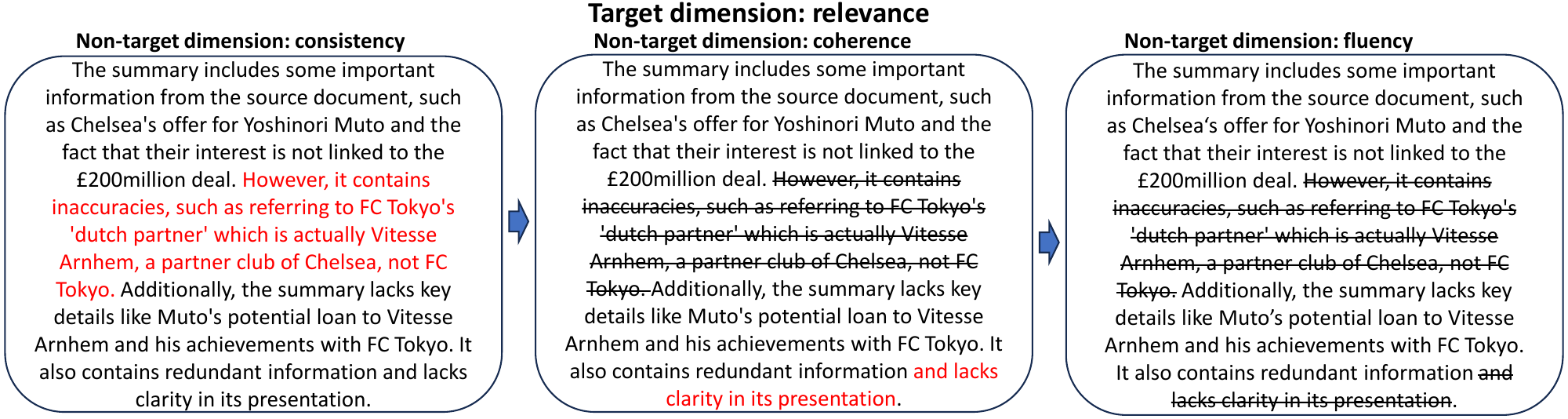}
\caption{Example workflow of DimCheck on a COT. The evidence which is unrelated to the target dimension and removed by Dimcheck in the current step is in \textcolor{red}{red}, evidence removed in all previous steps are with \sout{strikethrough}.}
\label{fig:dimcheck_example}
\end{figure*}

DimCheck is a method designed to reduce the inter-dimension dependence. DimCheck builds upon the Analyze-Rate framework \citep{chiang2023closer}, where an LLM judge first generates a COT $r_{i,t}$, and then generates a score $\hat{s}_{i,t}$ for the text $t$ on the target dimension $d_i$. To reduce inter-dimension dependence, DimCheck instruct an LLM to modify 
$r_{i,t}$ by removing evidence unrelated to the dimension $d_i$. An example modification process by DimCheck is in Fig. \ref{fig:dimcheck_example}. The LLM judge then assigns a new score based on the modified COT. 

Removing unrelated evidence without considering other dimensions is challenging, since the semantic boundaries of dimensions can be unclear and overlapping. Therefore, DimCheck iteratively removes evidence related to one non-target dimension $d' \in \{d_{k\neq i}\}$ at a time.  
For each iteration, the LLM is given evaluation 
The LLM is then instructed to 
identify and remove evidence in the COT $r_{i,t}$ that is 
related to $d'$ but not $d_i$ without changing other parts of the COT. 
If such evidence does not exist, the COT $r_{i,t}$ will remain unchanged. In rare cases, this process may remove the entire content of COT $r_{i,t}$. When this occurs, DimCheck discards the modification for that iteration, as Analyze-Rate requires a non-empty COT for scoring. To ensure accurate interpretation of evaluation dimensions and the COT $r_{i,t}$, DimCheck requires the model to think step-by-step before generating the modified COT. This process is repeated sequentially for each non-target dimension. The final modified COT is denoted as $r^{new}_{i,t}$, 
which only includes evidence related to the target dimension $d_i$. The prompt used by DimCheck is in App. \ref{app:prompt_dimcheck}. 

For texts whose COTs are modified, DimCheck then keeps the evaluation prompt unchanged and instructs the LLM judge to regenerate only the score $\hat{s}^{new}_{i,t}$ for text $t$ on dimension $d_i$ conditioned on the modified COT $r^{new}_{i,t}$. 
For texts whose COTs remain unchanged, DimCheck keeps the original score $\hat{s}_{i,t}$. DimCheck‘s pseudo code is in Alg. \ref{alg:dimcheck}.


\begin{algorithm}
\caption{Pseudo Code for DimCheck}\label{alg:dimcheck}
\centering
\scriptsize
\begin{algorithmic}[1]
\State\textbf{Input}: text supposed to be evaluated $t$; target dimension $d_i$; non-target dimensions $\{d_{k\neq i}\}$
\State $r_{i,t},\hat{s}_{i,t}=LLM_{judge}(t,d_i)$ \Comment{generate the COT $r_{i,t}$ and score $\hat{s}_{i,t}$ on dimension $d_i$} 
\State $r^{new}_{i,t}=r_{i,t}$; $\hat{s}^{new}_{i,t}$ = $\hat{s}_{i,t}$
\For {$d' \in \{d_{k\neq i}\}$} 
\State $r^{new}_{i,t}=EDIT(r^{new}_{i,t},d_i,d')$ \Comment{remove evidence related to $d'$ but unrelated to $d_i$ from the COT}
\EndFor 
\If {$r^{new}_{i,t} \neq r_{i,t}$}
\State $\hat{s}^{new}_{i,t}=LLM_{judge}(r^{new}_{i,t},t,d_i)$ \Comment{assign a new score if COT is modified}
\EndIf
\State \textbf{Return} $\hat{s}^{new}_{i,t}$ 
\end{algorithmic}
\end{algorithm}

\section{Experimental Setup and Dataset}

\begin{table}[]
\centering
\scriptsize
\setlength{\tabcolsep}{1mm}
\begin{tabular}{lcccc}
\toprule
             & Task                                                                         & \# Dim.                                                                                             & \#Inp. & \#Out.                                              \\ \midrule
SummEval     & news summarization                                                           & 4                          & 100    & 16                                                      \\
Topical-Chat & \begin{tabular}[c]{@{}c@{}} knowledge-grounded \\ conversation\end{tabular}  &  4                  & 60     & 6       \\
Hanna        & narrative generaion                                                             & 6      & 96     & 11                                                      \\
OpinSumm. & review summarization                                                        & 4 & 10     & 14                                                     \\ \bottomrule
\end{tabular}
\caption{
Datasets Overview. We show the task, number of evaluation dimensions (\# Dim.), number of inputs (\#Inp.) and number of outputs per input (\#Out.). }
\label{tab:dataset_stat}
\end{table}

The evaluation of inter-dimension dependence requires datasets with human annotations on multiple dimensions. We 
use following datasets that meet this requirement in diverse domains: (i) SummEval \citep{fabbri2020summeval} for news summarization; (ii) Topical-Chat \citep{gopalakrishnan2019topical} for knowledge-grounded conversation; (iii) Hanna \citep{chhun2024do} for narrative generation; (iv) OpinSummEval \citep{shen2023opinsummeval} for opinion summarization. The statistics are in Table \ref{tab:dataset_stat}. 

To evaluate inter-dimension dependence with CorrGap, we perform experiments on seven LLMs from four families:  Llama 3 \citep{llama3modelcard} (Llama3.1-8b-Instruct, Llama3.3-70b-Instruct), Qwen 3 \citep{qwen3technicalreport} (Qwen3-14B, Qwen3-32B), Gemma3 \citep{gemma_2025} (gemma-3-12b-it, gemma-3-27b-it), Ministral \citep{liu2026ministral} (Ministral-3-14B-Instruct-2512), GPT-5.4 \citep{singh2025openai} (gpt-5.4-nano, gpt-5.4-mini). We also consider an LLM dedicated for evaluation, M-Prometheus-14B \citep{pombalm}. During inference, we adopt top-p sampling with $p=0.9$ and temperature of $0.6$ for all LLMs.
To evaluate whether the correlation difference between groups is statistically significant, we use the permutation test \citep{pitman1937significance} 
by randomly splitting all texts into two groups $1000$ times 
while controlling the ground truth scores as in Sec.\ref{sec:corrgap}. CorrGap values are statistically significant when the correlation differences are statistically significant. 

To evaluate the performance of DimCheck, we perform experiments on Llama3.3-70b-Instruct, Qwen3-32B, and gemma-3-27b-it with the same inference hyperparameters as before. DimCheck uses the same LLM to modify the reasoning process. When removing unrelated evidence from the COT, DimCheck instructs Llama3.3-70b-Instruct and gemma-3-27b-it to first think step-by-step before modifying COTs. For Qwen3-32B, DimCheck enables the thinking mode. Implmentation details and computation cost analysis are in App. \ref{app:prompt_dimcheck}

\section{Experiment}

\subsection{Evaluation of CorrGap}
\label{sec:corrgap_effect}
\begin{table}[]
\centering
\scriptsize
\setlength{\tabcolsep}{1mm}
\begin{tabular}{lcccccc}
\toprule
                        & \multicolumn{2}{c}{Llama3.3-70b} & \multicolumn{2}{c}{Gemma3-27b} & \multicolumn{2}{c}{Qwen3-32b} \\
                        & Fact.       & Analyze.   & Fact.      & Analyze  & Fact.     & Analyze.  \\ \midrule
CorrGap                 & 5.9             & \textbf{20.5}  & 5.6            & \textbf{12.0} & 5.0           & \textbf{22.4} \\
~~ w/o control     & 6.2             & \textbf{24.6}  & \textbf{8.4}   & \textbf{20.7} & \textbf{11.5} & \textbf{24.9} \\
\bottomrule
\end{tabular}
\caption{Evaluation of CorrGap using FactScore (Fact.) and Analyze-Rate (Analyze.) on the SummEval dataset. The correlation differences that are statistically significant ($p<0.05$) are \textbf{bolded}. CorrGap correctly identify FactScore as having low inter-dimension dependence. }
\label{tab:factscore_eval} 
\end{table}

Evaluating CorrGap is challenging since it cannot be evaluated by just looking at each individual score and its COTs. Therefore, we evaluate CorrGap via two evaluations. 

The first evaluation considers an LLM judge especially constructed for a specific target dimension and hence by design, would be minimally influenced by other dimensions. If CorrGap is a good metric, it should correctly identify this judge as having insignificant inter-dimension dependence. 
An example of such a judge is FactScore \citep{min2023factscore}. It measures \textit{consistency} by decomposing summaries into atomic content units and computing the proportion of units entailed by the input document. Since the evaluation operates on atomic units rather than full sentences, it is unlikely to be affected by \textit{coherence} or \textit{fluency}. 
It is also less sensitive to \textit{relevance}, since it only checks whether each unit is supported by the document rather than whether it captures the main content. 
For comparison, we consider a more general-purpose LLM judge that is more likely to be influenced by other dimensions and use it for \textit{consistency} evaluation: Analyze-Rate. 
We compute CorrGap for both judges with \textit{consistency} as the target dimension on the SummEval dataset. 
We also compare an ablation of CorrGap that does not control ground truth scores during the split (CorrGap w/o control).
Results across Llama3.3-70b-Instruct, Qwen3-32B, and gemma-3-27b-it are in Tab.~\ref{tab:factscore_eval}.

The results show that CorrGap for FactScore is not statistically significant ($p<0.05$), whereas CorrGap for Analyze-Rate is statistically significant ($p>0.05$) across all three LLMs. This indicates that CorrGap correctly identifies FactScore as having low inter-dimension dependence. In contrast, CorrGap w/o control is statistically significant for FactScore with Gemma3-27B and Qwen3-32B. It shows that controlling the distribution of ground truth scores is necessary to avoid capturing spurious inter-dimension dependence.

\begin{table}[]
\centering
\scriptsize
\setlength{\tabcolsep}{1mm}
\begin{tabular}{lcccccccccc}
\toprule
 $\gamma$     & \multicolumn{2}{c}{SummEval} & \multicolumn{2}{c}{Topical Chat} & \multicolumn{2}{c}{Hanna}   & \multicolumn{2}{c}{OpinSumm.} & \multicolumn{2}{c}{Average} \\
      & CG       & w/o c. & CG          & w/o c.   & CG       & w/o c. & CG       & w/o c.   & CG       & w/o c. \\ \midrule
100 & \textbf{0.0}   & 9.1         & \textbf{0.0}     & 0.0           & \textbf{0.0}  & 0.0         & \textbf{0.5}  & 0.9           & \textbf{0.1}  & 2.5         \\
67  & \textbf{0.0}   & 0.1         & \textbf{0.0}     & 0.1           & \textbf{0.0}  & 0.0         & \textbf{5.0}  & 6.0           & \textbf{1.3}  & 1.6         \\
33  & \textbf{0.8}   & 2.3         & \textbf{3.5}     & 6.1           & \textbf{0.0}  & 0.1         & \textbf{14.6} & 18.9          & \textbf{4.7}  & 6.8         \\
0   & \textbf{23.8}  & 12.9        & \textbf{23.9}    & 20.0          & \textbf{23.2} & 21.3        & 24.3          & \textbf{24.8} & \textbf{23.8} & 19.8 \\   \bottomrule 
\end{tabular}
\caption{P-values of CorrGap (CG) and its variants without controlling the ground truth scores (w/o c.) on synthetic scores with different inter-dimension dependence ($\gamma$). The best-performing metric is \textbf{bolded}. CorrGap accurately distinguishes between synthetic scores with low or high inter-dimension dependence. }
\label{tab:noise_eval}
\end{table}

The second evaluation uses perturbed scores to test whether CorrGap can distinguish between scores with low or high inter-dimension dependence. We consider two perturbations. First, we perturb the ground truth score, $s_{i,t}$, with random noise, so that the perturbed scores $\hat{s}^r_{i,t}$ equals $s_{i,t}+1$, $s_{i,t}$, or $s_{i,t}-1$ with equal probability (33\% each). Second, we perturb ground truth score $s_{i,t}$, with inter-dimension dependence. The perturbed score $\hat{s}^d_{i,t}$ will be $s_{i,t}+1$ if the text performs better on other dimensions than the target dimension, i.e. $mean(\{s^p_{k\neq i, t}\})>s^p_{i,t}$, which uses percentile as in Sec. \ref{sec:corrgap}. Otherwise, the perturbed score $\hat{s}^d_{i,t}$ will be $s_{i,t}-1$. Perturbations that produce scores outside the valid range are discarded. Perturbed scores are then constructed by mixing $\gamma\%$ 
of ground truth scores with inter-dimension dependence $\hat{s}^d_i$ with $(100-\gamma)\%$ 
of ground truth scores with random noise $\hat{s}^r_i$. A good metric should show high p-values when $\gamma=0$, indicating no inter-dimension dependence, and small p-values when $\gamma>0$, indicating the existing inter-dimension dependence. We report the average p-values across all dimensions (in percentage) over 10 runs in Tab.~\ref{tab:noise_eval}. 

From the table, we observe that CorrGap is not statistically significant ($p>0.05$) when $\gamma=0$ for all datasets and statistically significant ($p<0.05$) when $\gamma$ is non-zero except for OpinSummEval. The results show that CorrGap accurately distinguishes between synthetic scores with low or high inter-dimension dependence. Moreover, compared with CorrGap w/o control, CorrGap shows higher p-values when $\gamma=0$ and lower p-values when $\gamma>0$. The results show that controlling the distribution of ground truth scores is essential for accurately measuring inter-dimension dependence. 

The correlation differences measured by CorrGap can be caused by inter-dimension dependence of human annotators who annotate the ground-truth scores instead of LLM judges. To check this, we compare inter-annotator agreement between the low- and high-variance groups. If human annotators were strongly affected by inter-dimension dependence, different annotators would likely be affected to different extents, resulting in much lower agreement in the high-variance group. Therefore, we report Randolph’s kappa\cite{randolph2005free} for both groups in Tab. \ref{tab:agreement_diff}. We find that inter-annotator agreement between two groups are similar, showing that human annotators are less likely to be affected by inter-dimension dependence. 

\begin{table}[]
\centering
\scriptsize
\setlength{\tabcolsep}{1mm}
\begin{tabular}{lcccc}
\toprule
              & SummEval & Topical Chat & Hanna & OpinSummEval \\ \midrule
Low Variance  & 50.43    & 43.84        & 8.33  & 83.36        \\
High Variance & 50.58    & 45.24        & 8.50  & 81.02   \\ \bottomrule    
\end{tabular}
\caption{Inter-annotator agreements for ground-truth scores in low- and high-variance groups. Similar inter-annotator agreements between two groups show that ground-truth scores are less likely to be affected by inter-dimension dependence.}
\label{tab:agreement_diff}
\end{table}

\subsection{Evaluating Inter-dimension Dependence with CorrGap}
\label{sec:corrgap_eval}
\begin{table*}[t]
\centering
\scriptsize
\begin{tabular}{lccccccccccc}
\toprule
                   & \multicolumn{2}{c}{SummEval[4]} & \multicolumn{2}{c}{Topical Chat[4]} & \multicolumn{2}{c}{Hanna[6]} & \multicolumn{2}{c}{OpinSumm.[4]} & \multicolumn{2}{c}{Avg.}                         \\
                   & $\tau\uparrow$             &  CG$\downarrow$         & $\tau\uparrow$               & CG$\downarrow$             & $\tau\uparrow$           & CG$\downarrow$          & $\tau\uparrow$               & CG$\downarrow$             & $\tau\uparrow$              & CG$\downarrow$             & overall$\uparrow$        \\ \midrule
              & \multicolumn{11}{c}{G-Eval \citep{liu2023g}}                                                \\
Llama3.1-8b   & 32.7 & 25.6(4) & 41.7 & 9.2(2)  & 25.6 & 9.6(2)  & 29.6 & 12.4(0) & 32.4 & 14.2 & 59.1 \\
Llama3.3-70b  & 25.9 & 15.4(4) & 34.1 & 11.6(2) & 28.3 & 11.4(4) & 33.9 & 23.6(3) & 30.5 & 15.5 & 57.5 \\
Gemma3-12b    & 39.1 & 22.4(4) & 45.4 & 17.5(3) & 18.0 & 7.1(2)  & 34.5 & 24.3(2) & 34.3 & 17.8 & 58.2 \\
Gemma3-27b    & 44.2 & 24.8(4) & 44.1 & 12.9(3) & 27.4 & 9.9(3)  & 37.2 & 25.6(2) & 38.2 & 18.3 & 60.0 \\
Qwen3-14b     & 40.7 & 25.8(4) & 46.3 & 18.6(3) & 26.5 & 12.0(5) & 37.9 & 24.8(2) & 37.9 & 20.3 & 58.8 \\
Qwen3-32b     & 39.4 & 24.7(4) & 47.8 & 13.2(3) & 32.0 & 11.1(4) & 33.6 & 17.9(2) & 38.2 & 16.7 & 60.7 \\
Ministral-14b & 39.2 & 24.7(4) & 49.8 & 13.2(3) & 26.8 & 18.7(5) & 34.5 & 28.7(2) & 37.6 & 21.3 & 58.1 \\ \hdashline
              & \multicolumn{11}{c}{Analyze-rate \citep{chiang2023closer}}                                          \\
Llama3.1-8b   & 30.1 & 23.9(4) & 32.5 & 10.0(1) & 22.8 & 7.8(3)  & 23.0 & 24.5(2) & 27.1 & 16.6 & 55.3 \\
Llama3.3-70b  & 46.1 & 30.0(4) & 53.3 & 13.1(3) & 34.6 & 11.3(3) & 32.0 & 18.9(3) & 41.5 & 18.3 & 61.6 \\
Gemma3-12b    & 42.8 & 25.8(4) & 46.3 & 9.0(1)  & 26.7 & 11.3(4) & 33.7 & 21.0(2) & 37.4 & 16.8 & 60.3 \\
Gemma3-27b    & 44.1 & 26.3(4) & 51.1 & 10.7(2) & 28.7 & 11.4(3) & 37.6 & 16.1(2) & 40.4 & 16.1 & 62.1 \\
Qwen3-14b     & 42.2 & 27.1(4) & 51.7 & 14.6(3) & 33.9 & 13.0(5) & 36.0 & 20.1(1) & 41.0 & 18.7 & 61.1 \\
Qwen3-32b     & 41.8 & 25.3(4) & 49.1 & 10.1(1) & 34.3 & 11.7(4) & 37.6 & 18.5(3) & 40.7 & 16.4 & 62.1 \\
Ministral-14b & 36.7 & 28.5(4) & 48.6 & 17.2(4) & 26.6 & 12.8(5) & 31.8 & 28.2(2) & 36.0 & 21.7 & 57.1 \\ 
GPT5.4-Nano & 37.7 & 22.6(4) & 38.8 & 9.7(1) & 29.5 & 8.9(3) & 43.0 & 21.8(2) & 37.2 & 15.7 & 60.8 \\
GPT5.4-Mini & 47.6 & 20.3(4) & 50.6 & 9.0(1) & 33.4 & 11.4(5) & 41.8 & 19.5(3) & 43.3 & 15.0 & 64.1 \\
M-Prom.-14b & 28.9 & 20.4(3) & 37.6 & 12.7(1) & 22.2 & 12.3(5) & 16.4 & 11.3(1) & 26.3 & 14.2 & 56.0 \\ \hdashline
              & \multicolumn{11}{c}{Debate \citep{kim2024debate}}                                                \\
Llama3.1-8b   & 22.8 & 18.8(4) & 29.6 & 9.6(1)  & 18.4 & 5.8(1)  & 15.0 & 13.4(0) & 21.4 & 11.9 & 54.8 \\
Llama3.3-70b  & 44.4 & 29.7(4) & 52.1 & 13.6(3) & 33.4 & 12.0(4) & 27.7 & 16.3(1) & 39.4 & 17.9 & 60.7 \\
Gemma3-12b    & 34.9 & 21.8(4) & 33.6 & 9.0(1)  & 19.3 & 11.0(4) & 19.4 & 22.6(2) & 26.8 & 16.1 & 55.3 \\
Gemma3-27b    & 31.0 & 18.0(4) & 52.0 & 11.7(2) & 25.1 & 12.3(5) & 16.1 & 12.1(0) & 31.0 & 13.5 & 58.8 \\
Qwen3-14b     & 40.6 & 27.0(4) & 52.7 & 13.6(3) & 34.0 & 12.8(5) & 35.0 & 23.2(1) & 40.6 & 19.1 & 60.7 \\
Qwen3-32b     & 41.8 & 23.5(4) & 49.9 & 10.3(2) & 30.8 & 11.9(4) & 30.5 & 18.9(2) & 38.2 & 16.2 & 61.0 \\
Ministral-14b & 19.1 & 12.7(3) & 23.7 & 11.9(0) & 22.0 & 11.1(4) & 18.8 & 21.2(1) & 20.9 & 14.2 & 53.3                  \\     \bottomrule         
\end{tabular}
\caption{Correlation ($\tau$), inter-dimension dependence ($CG$), and number of dimensions whose inter-dimension dependence is statistically significant (in brackets) for different evaluation frameworks and LLMs. Inter-dimension dependence is pervasive. }
\label{tab:corrgap_eval}
\end{table*}

In this section, we use CorrGap to measure the inter-dimension dependence of different LLM-judges based evaluation frameworks. We consider three evaluation frameworks: (i) G-Eval \citep{liu2023g}, which directly generates a score 
(ii) Analyze-rate \citep{chiang2023closer}, which first generates a COT before scoring; (iii) Debate \citep{kim2024debate}, which generates scores by a debate between a grader and a critic. We evaluate 10 LLMs across these frameworks. For M-Prom.-14B, we evaluate only Analyze-Rate, as it is tuned specifically for this framework. For GPT5.4, we also evaluate only Analyze-Rate because its output distribution is unavailable and Debate incurs high inference cost. Implementation details are in \ref{app:llm_judge_detail}. For each framework, we report the average CorrGap ($CG$) across all dimensions, along with the number of dimensions (in parentheses) with statistically significant CorrGap ($p<0.05$). To measure judge accuracy, we report the average Kendall's tau ($\tau$) between predicted and ground-truth scores across all dimensions. Following prior work \citep{liu2023g, chhun2024do, shen2023opinsummeval}, we use summary-level correlations for SummEval and OpinSummEval, and sample-level correlations for Topical Chat and Hanna. We also report an overall score, computed as the average of the correlation and one minus CorrGap across all datasets. Results are shown in Tab. \ref{tab:corrgap_eval}.

From the table, we observe that all evaluation frameworks and LLMs have dimensions whose CorrGap value are statistically significant, indicating that inter-dimension dependence is pervasive. 
Moreover, for all dimensions whose CorrGap values are statistically significant, the correlations in the high-variance group are consistently lower than those in the low-variance group, which meets our assumption in Sec. \ref{sec:corrgap}.  

Among all evaluation frameworks and LLMs, GPT5.4-Mini with Analyze-Rate shows the best overall performance. For smaller models (<=14b parameters), Qwen3-14b with Analyze-Rate shows the best overall performance. For comparison between evaluation frameworks, Analyze-rate generally shows the best overall performance except for Llama3.1-8b, while Debate and G-Eval show comparable performance among different LLMs. 
For comparison between LLMs, while larger LLMs generally show higher correlations than smaller LLMs, larger LLMs do not necessarily show lower CorrGap values than smaller models. The results show that when evaluating LLM judges, it is important to consider CorrGap in addition to correlation, as they show complimentary patterns. 

We also analyze whether inter-dimension dependence varies across texts of different quality levels. For each dimension, we split texts into low- and high-quality groups based on their ground-truth scores. We find that inter-dimension dependence is more severe for high-quality texts on SummEval and Hanna, but more severe for low-quality texts on Topical Chat. These results suggest that inter-dimension dependence varies across quality levels. Full results for each LLM are shown in App. \ref{app:corr_diff}.

We also analyze the CorrGap value for each dimension and find that inter-dimension dependence is more severe for some dimensions as in Table \ref{tab:dimension_analysis}.

\subsection{Evaluation of DimCheck}
\label{sec:eval_dimcheck}

\begin{table*}[t]
\centering
\scriptsize
\setlength{\tabcolsep}{1mm}
\begin{tabular}{lccccccccccc}
\toprule
                   & \multicolumn{2}{c}{SummEval[4]} & \multicolumn{2}{c}{Topical Chat[4]} & \multicolumn{2}{c}{Hanna[6]} & \multicolumn{2}{c}{OpinSumm.[4]} & \multicolumn{2}{c}{Avg.}                         \\
                  & $\tau\uparrow$             &  CG$\downarrow$         & $\tau\uparrow$               & CG$\downarrow$             & $\tau\uparrow$           & CG$\downarrow$          & $\tau\uparrow$               & CG$\downarrow$             & $\tau\uparrow$              & CG$\downarrow$             & overall$\uparrow$        \\ \midrule
                   & \multicolumn{11}{c}{Llama3.3-70b}                                                                                                                                                 \\ 
Analyze-Rate       & \textbf{46.1} & 30.0(4)          & 53.3          & 13.1(3)          & 34.6          & 11.3(3)          & 32.0          & 18.9(3)          & 41.5          & 18.3          & 61.6          \\
Analyze-Rate+Other & 45.1          & 30.5(4)          & 52.0          & 13.8(3)          & 35.9 & 12.6(3)          & \textbf{34.9} & 17.4(2)          & 42.0          & 18.6          & 61.7          \\
Analyze-Rate+Joint & 41.4 & 28.6(4) & 53.5 & 13.9(2) & \textbf{37.2} & 13.7(6) & 33.4 & 19.5(2) & 41.4 & 18.9 & 61.2 \\ 
Self-consistency      & 47.3 & 28.3 (4) & 52.7 & 14.3 (3) & 35.5 & 13.2 (4) & 34.6 & 19.3 (2) & 42.5 & 18.8 & 61.9 \\
Self-reflection & 46.1 & 29.3 (4) & 52.3 & 13.6 (3) & 32.5 & 12.9 (4) & 28.8 & 18.9 (2) & 39.9 & 18.7 & 60.6 \\
Checkeval                    & 43.3 & 25.3 (4) & 46.4 & 18.3 (2) & -   & -       & -   & -       & -   & -   & -   \\
Analyze-Rate+DimCheck     & 45.9          & \textbf{27.9}(4) & \textbf{54.1} & \textbf{12.5}(3) & 35.0          & \textbf{11.1}(3) & 33.2          & \textbf{14.1}(1) & \textbf{42.0} & \textbf{16.4} & \textbf{62.8}             \\
 \hdashline
                   & \multicolumn{11}{c}{Gemma3-27b}                                                                                                                                                                                                                                           \\ 
Analyze-Rate       & 44.1          & 26.3(4)          & 51.1          & 10.7(2)         & 28.7          & 11.4(3)         & 37.6          & 16.1(2)          & 40.4          & 16.1          & 62.1          \\
Analyze-Rate+Other & \textbf{45.6} & 27.0(4)          & 49.9          & 12.6(2)         & 29.8 & \textbf{8.5}(2) & \textbf{38.0} & 17.2(2)          & \textbf{40.8} & 16.3          & 62.3          \\
Analyze-Rate+Joint & 43.0 & 23.6(4) & 47.6 & 12.8(1) & \textbf{32.3} & 10.5(4) & 35.7 & \textbf{14.1}(1) & 39.6 & 15.2 & 62.2 \\
Self-consistency        & 46.7 & 28.1 (4) & 52.7 & 8.7 (1)  & 29.6 & 11.1 (5) & 38.7 & 19.1 (1) & 41.9 & 16.8 & 62.6 \\
Self-reflection & 33.0 & 23.9 (4) & 52.0 & 15.2 (3) & 27.7 & 12.3 (5) & 34.5 & 13.7 (3) & 36.8 & 16.3 & 60.3 \\
Checkeval                    & 38.8 & 22.5 (4) & 42.2 & 12.2 (2) & -   & -       & -   & -       & -   & -   & -   \\

Analyze-Rate+DimCheck     & 42.9          & \textbf{22.3}(4) & \textbf{51.3} & \textbf{9.5}(2) & 29.2          & 10.1(3)         & 38.0          & 15.0(1) & 40.3          & \textbf{14.2} & \textbf{63.1}             \\
 \hdashline

                   & \multicolumn{11}{c}{Qwen3-32b}                                                                                                                                                                                                                                                   \\
Analyze-Rate       & 41.8          & 25.3(4)          & 49.1          & 10.1(1)         & 34.3          & 11.7(4)         & \textbf{37.6} & 18.5(3)          & \textbf{40.7} & 16.4          & 62.1          \\
Analyze-Rate+Other & 41.0          & 27.6(4)          & 48.5          & 10.5(2)         & \textbf{35.6} & \textbf{9.4}(3) & 26.0          & 17.0(2)          & 37.8          & 16.1          & 60.8          \\
Analyze-Rate+Joint & \textbf{42.2} & 29.1(4)  & \textbf{52.0} & 17.2(4) & 34.1 & 14.2(5) & 35.4 & 21.7(2) & 40.9 & 20.5 & 60.2 \\
Self-consistency       & 43.4 & 27.3 (4) & 50.5 & 10.9 (2) & 35.8 & 12.2 (5) & 35.2 & 17.5 (1) & 41.3 & 17.0 & 62.1 \\
Self-reflection & 37.3 & 26.2 (4) & 50.1 & 10.6 (2) & 33.2 & 13.6 (3) & 33.0 & 12.7 (1) & 38.4 & 15.8 & 61.3 \\
Checkeval                    & 43.4 & 24.7 (4) & 42.1 & 14.0 (1) & -   & -       & -   & -       & -   & -   & -    \\
Analyze-Rate+DimCheck     & 42.0 & \textbf{23.9}(4) & 49.9 & \textbf{8.5}(1) & 33.7          & 10.9(4)         & 35.9          & \textbf{12.2}(1) & 40.4          & \textbf{13.9} & \textbf{63.3} \\ \bottomrule         
\end{tabular}

\caption{Correlation ($\tau$), inter-dimension dependence ($CG_\tau$), and number of dimensions whose inter-dimension dependence is statistically significant (in brackets) for methods on reducing inter-dimension dependence. The best-performing method is \textbf{bolded}. DimCheck shows the best overall performance.}
\label{tab:dimcheck_eval}
\end{table*}

In this section, we evaluate the performance of DimCheck on mitigating inter-dimension dependence. 
Our baselines are (i) original Analyze-Rate which shows the best overall score in Tab.\ref{tab:corrgap_eval}; (ii) Analyze-Rate with evaluation instructions that explicitly state that scores should not be influenced by any non-target dimensions (Analyze-Rate+Other); (iii) Analyze-Rate that jointly evaluates all dimensions at once instead of separately evaluating each dimension (Analyze-Rate+Joint) ; (iv) Self-consistency \cite{wangself}, which randomly samples multiple ($n=5$) COTs and scores and the final score is the majority of sampled scores; (v) Self-reflection \cite{madaan2023self}, which iteratively refine previous COTs and scores by checking whether it contains unrelated evidence; (vi) CheckEval \cite{lee-etal-2025-checkeval}, which evaluates each dimension using multiple Yes/No questions.   Implementation details of these baselines are in App. \ref{app:detail_baseline}. We report the results in Tab. \ref{tab:dimcheck_eval}. 

From the table, we observe that DimCheck consistently shows lower CorrGap values, comparable correlations with Analyze-Rate, and a higher overall score. Other baselines do not bring consistent improvements.  The difference in overall scores between DimCheck and the second-best performing method is statistically significant ($p<0.05$) using bootstrapping test. The results show that DimCheck can mitigate the inter-dimension dependence while retaining the accuracy of LLM judges. We show example edited COTs and proportions of COTs edited by DimCheck in App. \ref{app:cot_edit}. We show the performance of DimCheck is stable using different prompts in App. \ref{app:dimcheck_stable}. Besides, DimCheck shows the highest correlation on the high-variance group as in Tab. \ref{tab:corrgap_diff}. 

\subsection{Human Evaluation of DimCheck}
\label{sec:human_eval}
To evaluate whether DimCheck can accurately identify COTs that contain unrelated evidence, we perform a pairwise human evaluation. Each evaluation sample contains a pair of COTs that focus on the same target dimension: one identified by DimCheck as containing unrelated evidence and the other not. For each sample, human annotators are instructed to select which COTs contain less unrelated evidence. 
We then measure the accuracy of DimCheck as the proportion of samples where a majority of the human annotators agree with DimCheck’s identification. 
We randomly sample $10$ samples from each of the SummEval and Topical dataset and each of three LLMs used by DimCheck, resulting in $60$ samples in total. Each sample is annotated by three annotators from Amazon MTurk. Human evaluation's details are in App. \ref{app:human_eval}.

Among $60$ samples, the Randolph’s Kappa \citep{randolph2005free} between three annotators is $0.63$, showing a substantial correlation. The accuracy is $95$ percent for Llama3.3-70b, and $85$ percent for Gemma3-27b and Qwen3-32b. The results show that DimCheck can accurately identify COTs with unrelated evidence. 

\subsection{Ablation Study of DimCheck}
\label{sec:ablation}
\begin{table}[]
\centering
\scriptsize
\setlength{\tabcolsep}{1mm}
\begin{tabular}{lccccccccc}
\toprule
                       & \multicolumn{2}{c}{Llama3.3-70b} & \multicolumn{2}{c}{Gemma3-27b} & \multicolumn{2}{c}{Qwen3-32b} & \multicolumn{2}{c}{Avg.} &         \\
                       & $\tau\uparrow$       & CG$\downarrow$      & $\tau\uparrow$      & CG$\downarrow$     & $\tau\uparrow$     & CG$\downarrow$     & $\tau\uparrow$   & CG$\downarrow$ & overall \\ \midrule
DimCheck           & \textbf{42.0} & 16.4          & 40.3          & \textbf{14.2} & 40.4          & \textbf{13.9} & 40.9          & \textbf{14.8} & \textbf{52.7} \\
~~ w/1 step  & 42.0          & \textbf{16.3} & \textbf{40.4} & 15.6          & 40.3          & 15.6          & 40.9          & 15.8          & 52.2          \\
~~ w/o other & 42.0          & 16.9          & 40.4          & 15.9          & 40.8          & 15.8          & \textbf{41.1} & 16.2          & 52.1          \\
~~ w/o think & 41.7          & 16.6          & 40.1          & 15.5          & \textbf{40.8} & 15.1          & 40.9          & 15.7          & 52.3                   \\ \bottomrule
\end{tabular}
\caption{Average performance of DimCheck and its ablated variants across datasets.}
\label{tab:ablate_brief}
\end{table}

In this section, we compare DimCheck with the following ablations, which instruct LLMs to remove evidence unrelated to the target dimension (i) in a single step (w/1 step); (ii) without reference to any non-target dimensions (w/o other); and  (iii) without thinking (w/o think). We show overall results across datasets in Tab. \ref{tab:ablate_brief}.  Compared with ablations, DimCheck achieves the best overall performance, indicating that all design choices contribute to its effectiveness. Full results are shown in Tab. \ref{tab:ablation_eval}.

\subsection{Training Smaller LLMs for Efficient COT Editing}

\begin{table}[]
\centering
\scriptsize
\setlength{\tabcolsep}{0.8mm}
\begin{tabular}{lccccccccc}
\toprule
                       & \multicolumn{2}{c}{Llama3.3-70b} & \multicolumn{2}{c}{Gemma3-27b} & \multicolumn{2}{c}{Qwen3-32b} & \multicolumn{2}{c}{Avg.} &         \\
                       & $\tau\uparrow$       & CG$\downarrow$      & $\tau\uparrow$      & CG$\downarrow$     & $\tau\uparrow$     & CG$\downarrow$     & $\tau\uparrow$   & CG$\downarrow$ & overall \\ \midrule
Analyze-Rate           & 45.5              & 30.9         & 42.2             & 28.6        & 41.1            & 25.5        & 42.9          & 28.3     & 57.3    \\
~+DimCheck               & 45.1              & 28.5         & 41.4             & 24.3        & 40.7            & 22.9        & 42.4          & 25.2     & 58.6    \\
~~ w/1 step      & 43.8              & 28.9         & 41.2             & 26.4        & 41.2            & 24.8        & 42.1          & 26.7     & 57.7    \\
~~ w/o think     & 43.0              & 30.6         & 42.9             & 29.3        & 41.7            & 25.5        & 42.5          & 28.4     & 57.0    \\
~~ w/trained & 45.0              & 29.8         & 41.9             & 24.3        & 41.2            & 24.9        & 42.7          & 26.3     & 58.2 \\ \bottomrule
\end{tabular}

\caption{Performance of trained smaller LLMs to edit COTs (DimCheck w/trained). DimCheck w/trained shows close overall performance to DimCheck.}
\label{tab:train_eval}
\end{table}
DimCheck's design can lead to relatively high inference cost. To improve the efficiency, we train a smaller LLM to edit the COT in a single step. We perform experiments on the SummEval dataset since only this dataset's sample size is big enough for training and evaluation (50-50 split). We train smaller LLMs using SFT on the COTs generated and edited by larger LLMs within each LLM family. E.g., we perform SFT on Llama3.1-8b-Instruct using COTs from Llama3.3-70b-Instruct. All LLMs are trained for one epoch with a learning rate of 1e-4 and a batch size of 16. Implementation details are in App. \ref{app:sft_detail}. The performance of using trained smaller LLMs to edit COTs (DimCheck w/trained) and other baselines on the test set is in Tab. \ref{tab:train_eval}.

From the table, we observe that DimCheck w/trained shows close overall performance to DimCheck. It also outperforms DimCheck w/1 step and DimCheck w/o think, which edits either COTs in a single step or without thinking. 
The results show that a smaller trained LLM can achieve a performance close to that of a larger LLM with reduced inference costs. 

\section{Conclusion}
To quantify inter-dimension dependence of LLM judges, we propose CorrGap based on correlation differences accross groups. We show that CorrGap can reliably measure inter-dimension dependence and inter-dimension dependence is pervasive across nine LLMs and four tasks. To mitigate inter-dimension dependence, we propose DimCheck by iteratively removing unrelated evidence from the COTs of LLM judges. Our experiments show that DimCheck mitigates inter-dimension dependence and outperforms strong baselines across three LLMs and four tasks.

\section{Limitation}
This work focuses on pointwise evaluation, where the LLM judges assign a score to represent the quality of the generated text. While the underlying notion of inter-dimension dependence may also arise in other evaluation paradigms, such as pairwise preference judgments or listwise evaluation, these settings involve different evaluation frameworks and are beyond the scope of this work. Extending the analysis to broader evaluation paradigms is an interesting direction for future research. 

Besides, DimCheck is designed for LLM judges that explicitly generate a COT before assigning a score. While this evaluation frameworks are widely used, DimCheck is not directly applicable to methods that do not generate COTs. Extending the approach to these settings is an interesting direction for future work.

\section{Ethics Statement}
The datasets we use are all publicly available.  All the models used in this paper are publicly accessible. The inference and finetuning of models are performed on four Nvidia A6000 or Nvidia L40 GPUs. We do not annotate any data on our own.

We perform human evaluation experiments on Amazon Mechanical Turk. The annotators were compensated at a rate of \$20 per hour. During the evaluation, human annotators were not exposed to any sensitive or explicit content.

We use LLMs to polish the writing of the paper. 
\bibliography{custom}

\begin{thebibliography}{42}
\providecommand{\natexlab}[1]{#1}

\bibitem[{AI@Meta(2024)}]{llama3modelcard}
AI@Meta. 2024.
\newblock \href {https://github.com/meta-llama/llama3/blob/main/MODEL_CARD.md} {Llama 3 model card}.

\bibitem[{Bao et~al.(2026)Bao, Huang, Wang, Zhang, Zhou, Yang, Zhang, and Ye}]{bao2026position}
Han Bao, Yue Huang, Xiaoda Wang, Zheyuan Zhang, Yujun Zhou, Carl Yang, Xiangliang Zhang, and Yanfang Ye. 2026.
\newblock Position: General alignment has hit a ceiling; edge alignment must be taken seriously.
\newblock \emph{arXiv preprint arXiv:2602.20042}.

\bibitem[{Chan et~al.(2024)Chan, Chen, Su, Yu, Xue, Zhang, Fu, and Liu}]{chanchateval}
Chi-Min Chan, Weize Chen, Yusheng Su, Jianxuan Yu, Wei Xue, Shanghang Zhang, Jie Fu, and Zhiyuan Liu. 2024.
\newblock Chateval: Towards better llm-based evaluators through multi-agent debate.
\newblock In \emph{The Twelfth International Conference on Learning Representations}.

\bibitem[{Chhun et~al.(2024)Chhun, Suchanek, and Clavel}]{chhun2024do}
Cyril Chhun, Fabian~M. Suchanek, and Chlo{\'e} Clavel. 2024.
\newblock \href {https://doi.org/10.1162/tacl_a_00689} {Do language models enjoy their own stories? {P}rompting large language models for automatic story evaluation}.
\newblock \emph{Transactions of the Association for Computational Linguistics}, 12:1122--1142.

\bibitem[{Chiang and Lee(2023)}]{chiang2023closer}
Cheng-Han Chiang and Hung-yi Lee. 2023.
\newblock A closer look into using large language models for automatic evaluation.
\newblock In \emph{Findings of the Association for Computational Linguistics: EMNLP 2023}, pages 8928--8942.

\bibitem[{Cochran(1954)}]{cochran1954combination}
William~G Cochran. 1954.
\newblock The combination of estimates from different experiments.
\newblock \emph{Biometrics}, 10(1):101--129.

\bibitem[{Fabbri et~al.(2020)Fabbri, Kry{\'s}ci{\'n}ski, McCann, Xiong, Socher, and Radev}]{fabbri2020summeval}
Alexander~R Fabbri, Wojciech Kry{\'s}ci{\'n}ski, Bryan McCann, Caiming Xiong, Richard Socher, and Dragomir Radev. 2020.
\newblock Summeval: Re-evaluating summarization evaluation.
\newblock \emph{arXiv preprint arXiv:2007.12626}.

\bibitem[{Fu et~al.(2024)Fu, Ng, Jiang, and Liu}]{fu2024gptscore}
Jinlan Fu, See~Kiong Ng, Zhengbao Jiang, and Pengfei Liu. 2024.
\newblock Gptscore: Evaluate as you desire.
\newblock In \emph{Proceedings of the 2024 Conference of the North American Chapter of the Association for Computational Linguistics: Human Language Technologies (Volume 1: Long Papers)}, pages 6556--6576.

\bibitem[{Gao et~al.(2023)Gao, Ruan, Sun, Yin, Yang, and Wan}]{gao2023human}
Mingqi Gao, Jie Ruan, Renliang Sun, Xunjian Yin, Shiping Yang, and Xiaojun Wan. 2023.
\newblock Human-like summarization evaluation with chatgpt.
\newblock \emph{arXiv preprint arXiv:2304.02554}.

\bibitem[{Gopalakrishnan et~al.(2019)Gopalakrishnan, Hedayatnia, Chen, Gottardi, Kwatra, Venkatesh, Gabriel, and Hakkani-Tür}]{gopalakrishnan2019topical}
Karthik Gopalakrishnan, Behnam Hedayatnia, Qinlang Chen, Anna Gottardi, Sanjeev Kwatra, Anu Venkatesh, Raefer Gabriel, and Dilek Hakkani-Tür. 2019.
\newblock \href {https://doi.org/10.21437/Interspeech.2019-3079} {{Topical-Chat: Towards Knowledge-Grounded Open-Domain Conversations}}.
\newblock In \emph{Proc. Interspeech 2019}, pages 1891--1895.

\bibitem[{Goyal et~al.(2022)Goyal, Li, and Durrett}]{goyal2022news}
Tanya Goyal, Junyi~Jessy Li, and Greg Durrett. 2022.
\newblock News summarization and evaluation in the era of gpt-3.
\newblock \emph{arXiv preprint arXiv:2209.12356}.

\bibitem[{Gunjal et~al.(2025)Gunjal, Wang, Lau, Nath, He, Liu, and Hendryx}]{gunjal2025rubrics}
Anisha Gunjal, Anthony Wang, Elaine Lau, Vaskar Nath, Yunzhong He, Bing Liu, and Sean~M Hendryx. 2025.
\newblock Rubrics as rewards: Reinforcement learning beyond verifiable domains.
\newblock In \emph{NeurIPS 2025 Workshop on Efficient Reasoning}.

\bibitem[{He and Maghsudi(2025)}]{he2025pareto}
Qiang He and Setareh Maghsudi. 2025.
\newblock Pareto multi-objective alignment for language models.
\newblock In \emph{Joint European Conference on Machine Learning and Knowledge Discovery in Databases}, pages 257--272. Springer.

\bibitem[{Hosking et~al.(2024)Hosking, Blunsom, and Bartolo}]{hoskinghuman}
Tom Hosking, Phil Blunsom, and Max Bartolo. 2024.
\newblock Human feedback is not gold standard.
\newblock In \emph{The Twelfth International Conference on Learning Representations}.

\bibitem[{Hu et~al.(2022)Hu, Shen, Wallis, Allen-Zhu, Li, Wang, Wang, Chen et~al.}]{hu2022lora}
Edward~J Hu, Yelong Shen, Phillip Wallis, Zeyuan Allen-Zhu, Yuanzhi Li, Shean Wang, Liang Wang, Weizhu Chen, et~al. 2022.
\newblock Lora: Low-rank adaptation of large language models.
\newblock \emph{Iclr}, 1(2):3.

\bibitem[{Huang et~al.(2025)Huang, Zhuang, Lu, Qin, Xu, Zhao, Peng, Hu, Shen, Hu et~al.}]{huang2025reinforcement}
Zenan Huang, Yihong Zhuang, Guoshan Lu, Zeyu Qin, Haokai Xu, Tianyu Zhao, Ru~Peng, Jiaqi Hu, Zhanming Shen, Xiaomeng Hu, et~al. 2025.
\newblock Reinforcement learning with rubric anchors.
\newblock \emph{arXiv preprint arXiv:2508.12790}.

\bibitem[{Kim et~al.(2024{\natexlab{a}})Kim, Kim, and Yoon}]{kim2024debate}
Alex Kim, Keonwoo Kim, and Sangwon Yoon. 2024{\natexlab{a}}.
\newblock Debate: Devil’s advocate-based assessment and text evaluation.
\newblock In \emph{Findings of the Association for Computational Linguistics ACL 2024}, pages 1885--1897.

\bibitem[{Kim et~al.(2023)Kim, Shin, Cho, Jang, Longpre, Lee, Yun, Shin, Kim, Thorne et~al.}]{kim2023prometheus}
Seungone Kim, Jamin Shin, Yejin Cho, Joel Jang, Shayne Longpre, Hwaran Lee, Sangdoo Yun, Seongjin Shin, Sungdong Kim, James Thorne, et~al. 2023.
\newblock Prometheus: Inducing fine-grained evaluation capability in language models.
\newblock In \emph{The Twelfth International Conference on Learning Representations}.

\bibitem[{Kim et~al.(2024{\natexlab{b}})Kim, Suk, Longpre, Lin, Shin, Welleck, Neubig, Lee, Lee, and Seo}]{kim-etal-2024-prometheus}
Seungone Kim, Juyoung Suk, Shayne Longpre, Bill~Yuchen Lin, Jamin Shin, Sean Welleck, Graham Neubig, Moontae Lee, Kyungjae Lee, and Minjoon Seo. 2024{\natexlab{b}}.
\newblock \href {https://doi.org/10.18653/v1/2024.emnlp-main.248} {Prometheus 2: An open source language model specialized in evaluating other language models}.
\newblock In \emph{Proceedings of the 2024 Conference on Empirical Methods in Natural Language Processing}, pages 4334--4353, Miami, Florida, USA. Association for Computational Linguistics.

\bibitem[{Kumar et~al.(2025)Kumar, Nargund, and Sridhar}]{kumar2025courteval}
Sandeep Kumar, Abhijit~A Nargund, and Vivek Sridhar. 2025.
\newblock Courteval: A courtroom-based multi-agent evaluation framework.
\newblock In \emph{Findings of the Association for Computational Linguistics: ACL 2025}, pages 25875--25887.

\bibitem[{Lambert et~al.(2025)Lambert, Pyatkin, Morrison, Miranda, Lin, Chandu, Dziri, Kumar, Zick, Choi et~al.}]{lambert2025rewardbench}
Nathan Lambert, Valentina Pyatkin, Jacob Morrison, Lester James~Validad Miranda, Bill~Yuchen Lin, Khyathi Chandu, Nouha Dziri, Sachin Kumar, Tom Zick, Yejin Choi, et~al. 2025.
\newblock Rewardbench: Evaluating reward models for language modeling.
\newblock In \emph{Findings of the Association for Computational Linguistics: NAACL 2025}, pages 1755--1797.

\bibitem[{Lee et~al.(2025{\natexlab{a}})Lee, Kim, Kim, Cho, Kang, Kang, and Kim}]{lee2025checkeval}
Yukyung Lee, Joonghoon Kim, Jaehee Kim, Hyowon Cho, Jaewook Kang, Pilsung Kang, and Najoung Kim. 2025{\natexlab{a}}.
\newblock Checkeval: A reliable llm-as-a-judge framework for evaluating text generation using checklists.
\newblock In \emph{Proceedings of the 2025 Conference on Empirical Methods in Natural Language Processing}, pages 15782--15809.

\bibitem[{Lee et~al.(2025{\natexlab{b}})Lee, Kim, Kim, Cho, Kang, Kang, and Kim}]{lee-etal-2025-checkeval}
Yukyung Lee, JoongHoon Kim, Jaehee Kim, Hyowon Cho, Jaewook Kang, Pilsung Kang, and Najoung Kim. 2025{\natexlab{b}}.
\newblock \href {https://doi.org/10.18653/v1/2025.emnlp-main.796} {{C}heck{E}val: A reliable {LLM}-as-a-judge framework for evaluating text generation using checklists}.
\newblock In \emph{Proceedings of the 2025 Conference on Empirical Methods in Natural Language Processing}, pages 15771--15798, Suzhou, China. Association for Computational Linguistics.

\bibitem[{Liu et~al.(2026)Liu, Khandelwal, Subramanian, Jouault, Rastogi, Sad{\'e}, Jeffares, Jiang, Cahill, Gavaudan et~al.}]{liu2026ministral}
Alexander~H Liu, Kartik Khandelwal, Sandeep Subramanian, Victor Jouault, Abhinav Rastogi, Adrien Sad{\'e}, Alan Jeffares, Albert Jiang, Alexandre Cahill, Alexandre Gavaudan, et~al. 2026.
\newblock Ministral 3.
\newblock \emph{arXiv preprint arXiv:2601.08584}.

\bibitem[{Liu et~al.(2023)Liu, Iter, Xu, Wang, Xu, and Zhu}]{liu2023g}
Yang Liu, Dan Iter, Yichong Xu, Shuohang Wang, Ruochen Xu, and Chenguang Zhu. 2023.
\newblock G-eval: Nlg evaluation using gpt-4 with better human alignment.
\newblock In \emph{Proceedings of the 2023 Conference on Empirical Methods in Natural Language Processing}, pages 2511--2522.

\bibitem[{Madaan et~al.(2023)Madaan, Tandon, Gupta, Hallinan, Gao, Wiegreffe, Alon, Dziri, Prabhumoye, Yang et~al.}]{madaan2023self}
Aman Madaan, Niket Tandon, Prakhar Gupta, Skyler Hallinan, Luyu Gao, Sarah Wiegreffe, Uri Alon, Nouha Dziri, Shrimai Prabhumoye, Yiming Yang, et~al. 2023.
\newblock Self-refine: Iterative refinement with self-feedback.
\newblock \emph{Advances in neural information processing systems}, 36:46534--46594.

\bibitem[{Min et~al.(2023)Min, Krishna, Lyu, Lewis, Yih, Koh, Iyyer, Zettlemoyer, and Hajishirzi}]{min2023factscore}
Sewon Min, Kalpesh Krishna, Xinxi Lyu, Mike Lewis, Wen-tau Yih, Pang Koh, Mohit Iyyer, Luke Zettlemoyer, and Hannaneh Hajishirzi. 2023.
\newblock Factscore: Fine-grained atomic evaluation of factual precision in long form text generation.
\newblock In \emph{Proceedings of the 2023 Conference on Empirical Methods in Natural Language Processing}, pages 12076--12100.

\bibitem[{Pitman(1937)}]{pitman1937significance}
Edwin~JG Pitman. 1937.
\newblock Significance tests which may be applied to samples from any populations.
\newblock \emph{Supplement to the Journal of the Royal Statistical Society}, 4(1):119--130.

\bibitem[{Pombal et~al.(2025)Pombal, Yoon, Fernandes, Wu, Kim, Rei, Neubig, and Martins}]{pombalm}
Jos{\'e} Pombal, Dongkeun Yoon, Patrick Fernandes, Ian Wu, Seungone Kim, Ricardo Rei, Graham Neubig, and Andre Martins. 2025.
\newblock M-prometheus: A suite of open multilingual llm judges.
\newblock In \emph{Second Conference on Language Modeling}.

\bibitem[{Randolph(2005)}]{randolph2005free}
Justus~J Randolph. 2005.
\newblock Free-marginal multirater kappa (multirater k [free]): An alternative to fleiss' fixed-marginal multirater kappa.
\newblock \emph{Online submission}.

\bibitem[{Shen and Wan(2023)}]{shen2023opinsummeval}
Yuchen Shen and Xiaojun Wan. 2023.
\newblock Opinsummeval: Revisiting automated evaluation for opinion summarization.
\newblock \emph{arXiv preprint arXiv:2310.18122}.

\bibitem[{Singh et~al.(2025)Singh, Fry, Perelman, Tart, Ganesh, El-Kishky, McLaughlin, Low, Ostrow, Ananthram et~al.}]{singh2025openai}
Aaditya Singh, Adam Fry, Adam Perelman, Adam Tart, Adi Ganesh, Ahmed El-Kishky, Aidan McLaughlin, Aiden Low, AJ~Ostrow, Akhila Ananthram, et~al. 2025.
\newblock Openai gpt-5 system card.
\newblock \emph{arXiv preprint arXiv:2601.03267}.

\bibitem[{Team(2025{\natexlab{a}})}]{gemma_2025}
Gemma Team. 2025{\natexlab{a}}.
\newblock \href {https://goo.gle/Gemma3Report} {Gemma 3}.

\bibitem[{Team(2025{\natexlab{b}})}]{qwen3technicalreport}
Qwen Team. 2025{\natexlab{b}}.
\newblock \href {https://arxiv.org/abs/2505.09388} {Qwen3 technical report}.
\newblock \emph{Preprint}, arXiv:2505.09388.

\bibitem[{Wang et~al.(2023{\natexlab{a}})Wang, Liang, Meng, Sun, Shi, Li, Xu, Qu, and Zhou}]{wang2023chatgpt}
Jiaan Wang, Yunlong Liang, Fandong Meng, Zengkui Sun, Haoxiang Shi, Zhixu Li, Jinan Xu, Jianfeng Qu, and Jie Zhou. 2023{\natexlab{a}}.
\newblock Is chatgpt a good nlg evaluator? a preliminary study.
\newblock In \emph{Proceedings of the 4th New Frontiers in Summarization Workshop}, pages 1--11.

\bibitem[{Wang et~al.(2023{\natexlab{b}})Wang, Wei, Schuurmans, Le, Chi, Narang, Chowdhery, and Zhou}]{wangself}
Xuezhi Wang, Jason Wei, Dale Schuurmans, Quoc~V Le, Ed~H Chi, Sharan Narang, Aakanksha Chowdhery, and Denny Zhou. 2023{\natexlab{b}}.
\newblock Self-consistency improves chain of thought reasoning in language models.
\newblock In \emph{The Eleventh International Conference on Learning Representations}.

\bibitem[{Wei et~al.(2022)Wei, Wang, Schuurmans, Bosma, Xia, Chi, Le, Zhou et~al.}]{wei2022chain}
Jason Wei, Xuezhi Wang, Dale Schuurmans, Maarten Bosma, Fei Xia, Ed~Chi, Quoc~V Le, Denny Zhou, et~al. 2022.
\newblock Chain-of-thought prompting elicits reasoning in large language models.
\newblock \emph{Advances in neural information processing systems}, 35:24824--24837.

\bibitem[{Wu et~al.(2025)Wu, Hossain, Wood, Akbar, Chin, and Cornejo}]{wu-etal-2025-seeval}
Meng-Chen Wu, Md~Mosharaf Hossain, Tess Wood, Shayan~Ali Akbar, Si-Chi Chin, and Erwin Cornejo. 2025.
\newblock \href {https://doi.org/10.18653/v1/2025.findings-naacl.411} {{SEE}val: Advancing {LLM} text evaluation efficiency and accuracy through self-explanation prompting}.
\newblock In \emph{Findings of the Association for Computational Linguistics: NAACL 2025}, pages 7372--7383, Albuquerque, New Mexico. Association for Computational Linguistics.

\bibitem[{Xiao et~al.(2023)Xiao, Zhang, Lai, and Liao}]{xiao2023evaluating}
Ziang Xiao, Susu Zhang, Vivian Lai, and Q~Vera Liao. 2023.
\newblock Evaluating evaluation metrics: A framework for analyzing nlg evaluation metrics using measurement theory.
\newblock In \emph{Proceedings of the 2023 Conference on Empirical Methods in Natural Language Processing}, pages 10967--10982.

\bibitem[{Yang et~al.(2024)Yang, Pan, Luo, Qiu, Zhong, Yu, and Chen}]{yang2024rewards}
Rui Yang, Xiaoman Pan, Feng Luo, Shuang Qiu, Han Zhong, Dong Yu, and Jianshu Chen. 2024.
\newblock Rewards-in-context: Multi-objective alignment of foundation models with dynamic preference adjustment.
\newblock In \emph{International Conference on Machine Learning}, pages 56276--56297. PMLR.

\bibitem[{Ye et~al.(2025)Ye, Wang, Huang, Chen, Zhang, Moniz, Gao, Geyer, Huang, Chen et~al.}]{ye2025justice}
Jiayi Ye, Yanbo Wang, Yue Huang, Dongping Chen, Qihui Zhang, Nuno Moniz, Tian Gao, Werner Geyer, Chao Huang, Pin-Yu Chen, et~al. 2025.
\newblock Justice or prejudice? quantifying biases in llm-as-a-judge.
\newblock In \emph{International Conference on Learning Representations}.

\bibitem[{Zheng et~al.(2023)Zheng, Chiang, Sheng, Zhuang, Wu, Zhuang, Lin, Li, Li, Xing et~al.}]{zheng2023judging}
Lianmin Zheng, Wei-Lin Chiang, Ying Sheng, Siyuan Zhuang, Zhanghao Wu, Yonghao Zhuang, Zi~Lin, Zhuohan Li, Dacheng Li, Eric Xing, et~al. 2023.
\newblock Judging llm-as-a-judge with mt-bench and chatbot arena.
\newblock \emph{Advances in neural information processing systems}, 36:46595--46623.

\end{thebibliography}
\appendix

\section{Appendix}
\begin{table}[]
\scriptsize
\setlength{\tabcolsep}{0.5mm}
\centering
\begin{tabular}{lccccc}
\toprule
                      & SummEval & Topi. Chat & Hanna & OpinSumm. & Avg.          \\ \midrule
\multicolumn{6}{c}{Llama3.3-70b}                                                              \\
Analyze-Rate          & 29.9     & 47.1         & 29.0  & 25.4         & 32.8          \\
Analyze-Rate+Other    & 28.6     & 45.3         & 29.6  & 29.1         & 33.2          \\
Analyze-Rate+Joint    & 25.9     & 46.8         & 30.4  & 26.7         & 32.5          \\
Self-consistency      & 31.3     & 46.0         & 28.9  & 26.9         & 33.3          \\
Self-reflection       & 29.5     & 45.8         & 26.1  & 23.3         & 31.2          \\
CheckEval             & 27.0     & 37.9         &       &              &           \\
Analyze-Rate+DimCheck & 30.6     & 48.1         & 29.5  & 28.7         & \textbf{34.2} \\ \hdashline
\multicolumn{6}{c}{Gemma3-27b}                                                              \\
Analyze-Rate          & 29.2     & 46.0         & 24.0  & 29.0         & 32.0          \\
Analyze-Rate+Other    & 29.6     & 43.9         & 26.1  & 29.3         & 32.2          \\
Analyze-Rate+Joint    & 29.2     & 43.3         & 27.9  & 28.0         & 32.1          \\
Self-consistency      & 29.3     & 47.5         & 24.9  & 29.2         & 32.7          \\
Self-reflection       & 19.6     & 44.6         & 21.4  & 27.7         & 28.3          \\
CheckEval             & 22.9     & 36.5         &       &              &          \\
Analyze-Rate+DimCheck & 30.0     & 46.8         & 25.2  & 29.6         & \textbf{32.9} \\ \hdashline
\multicolumn{6}{c}{Qwen3-32b}                                                               \\
Analyze-Rate          & 28.4     & 45.4         & 28.5  & 27.8         & 32.5          \\
Analyze-Rate+Other    & 26.3     & 43.3         & 30.9  & 19.9         & 30.1          \\
Analyze-Rate+Joint    & 26.4     & 43.5         & 27.0  & 26.6         & 30.9          \\
Self-consistency      & 28.5     & 45.3         & 29.7  & 27.6         & 32.8          \\
Self-reflection       & 23.2     & 44.9         & 26.5  & 29.2         & 31.0          \\
CheckEval             & 28.2     & 35.0         &       &              &           \\
Analyze-Rate+DimCheck & 30.3     & 46.5         & 28.2  & 30.8         & \textbf{33.9} \\ \bottomrule
\end{tabular}
\caption{Correlation between predicted and ground-truth scores on high-variance groups for different methods. The best average correlation across datasets is \textbf{bolded}. DimCheck shows the highest average correlation among all methods. }
\label{tab:corr_lowgroup}
\end{table}
\subsection{Implementation Details of DimCheck}
\label{app:prompt_dimcheck}
\begin{table*}[]
\centering
\scriptsize
\begin{tabular}{p{15cm}}
You will be given a reasoning process generated by an LLM judge that is instructed to evaluate a \{output\_type\} on a target dimension. You will also be given evaluation instructions on the target dimension and a non-target dimension.                                                                                                                                                                                                    \\
Your task is to first check the reasoning process to find out whether it contains evidence that is closely related to the non-target dimension but not related to the target dimension. For this step, you should not consider any dimensions other than the given target dimension and non-target dimension.                                                                                                                               \\
If the reasoning process contains such evidence, you should remove all such evidence from the reasoning process. You should only make necessary modifications without changing other parts of the reasoning process. You should not add content that is not already presented in the original reasoning process. You should then return the modified reasoning process without extra comments. If all contents are removed, return 'EMPTY'. \\
If you do not find such evidence, return 'NO CHANGE'.                                                                                                                                                                                                                                                                                                                                                                                       \\
Reasoning Process:                                                                                                                                                                                                                                                                                                                                                                                                                          \\
\{reasoning\_process\}                                                                                                                                                                                                                                                                                                                                                                                                                         \\
Evaluation instruction for the target dimension:                                                                                                                                                                                                                                                                                                                                                                                            \\
\{evaluation\_instruction\_target\}                                                                                                                                                                                                                                                                                                                                                                                                             \\
Evaluation instruction for the non-target dimension:                                                                                                                                                                                                                                                                                                                                                                                        \\
\{evaluation\_instruction\_non\_target\}                                                                                                                                                                                                                                                                                                                                                                                                         \\
You should first think step-by-step and then return in the following format:                                                                                                                                                                                                                                                                                                                                                                \\
THINK: ...                                                                                                                                                                                                                                                                                                                                                                                                                                  \\
ANSWER: modified reasoning process or 'EMPTY' or 'NO CHANGE'                                                                                                                                                                                                                                                                                                                                                                  
\end{tabular}
\caption{Prompt used by DimCheck}
\label{tab:prompt_dimcheck}
\end{table*}

We show the prompt used by DimCheck in Tab. \ref{tab:prompt_dimcheck}. In the process of DimCheck, we find LLMs can think endlessly without generating answers in some rare cases. To mitigate this, we cap the maximum generation length at $2000$ tokens. If the model fails to produce an answer within this limit, DimCheck will restart the generation. 

In our experiments, non-target dimensions are removed sequentially following the order in which dimensions are introduced in the corresponding dataset papers. Specifically, for the SummEval dataset, the order is 'relevance', 'consistency’, 'coherence', and 'fluency'. For the Topical Chat dataset, the order is 'engaging', 'natural', 'coherence', and 'groundness'. For the Hanna dataset, the order is 'relevance', 'coherence', 'empathy', 'surprise', 'engagement', and 'complexity'. For the OpinSummEval dataset, the order is 'aspect relevance', 'self coherence', 'sentiment consistency', and 'readability'. To test whether DimCheck is order-sensitive, we evaluate DimCheck with Llama-3.3-70b using the random order (w/rand). We report the results in the Tab. \ref{tab:dimcheck_order}. 
\begin{table*}[]
\centering
\scriptsize
\setlength{\tabcolsep}{1mm}
\begin{tabular}{lccccccccccc}
\toprule
                   & \multicolumn{2}{c}{SummEval[4]} & \multicolumn{2}{c}{Topical Chat[4]} & \multicolumn{2}{c}{Hanna[6]} & \multicolumn{2}{c}{OpinSumm.[4]} & \multicolumn{2}{c}{Avg.}                         \\
                  & $\tau\uparrow$             &  CG$\downarrow$         & $\tau\uparrow$               & CG$\downarrow$             & $\tau\uparrow$           & CG$\downarrow$          & $\tau\uparrow$               & CG$\downarrow$             & $\tau\uparrow$              & CG$\downarrow$             & overall$\uparrow$        \\ \midrule
                   & \multicolumn{11}{c}{Llama3.3-70b}                                                                                                                                                 \\ 
Analyze-Rate+DimCheck & 45.9 & 27.9(4) & 54.1 & 12.5(3) & 35   & 11.1(3) & 33.2 & 14.1(1) & 42   & 16.4 & 62.8 \\
~~w/rand   & 45.2 & 27.3(4) & 54.7 & 11.5(2) & 35.2 & 10.9(3) & 33.6 & 14.9(1) & 42.2 & 16.1 & 63.0 \\
 \hdashline
                   & \multicolumn{11}{c}{Gemma3-27b}                                                                                                                                                                                                                                           \\ 
Analyze-Rate+DimCheck & 42.9 & 22.3(4) & 51.3 & 9.5(2)  & 29.2 & 10.1(3) & 38   & 15.0(1) & 40.3 & 14.2 & 63.1 \\
~~w/rand   & 42.1 & 19.9(4) & 51.1 & 10.5(2) & 29.1 & 10.9(3) & 37.5 & 15.1(1) & 39.9 & 14.1 & 62.9 \\
 \hdashline

                   & \multicolumn{11}{c}{Qwen3-32b}                                                                                                                                                                                                                                                   \\
Analyze-Rate+DimCheck & 42.0 & 23.9(4) & 49.9 & 8.5(1) & 33.7 & 10.9(4) & 35.9 & 12.2(1) & 40.4 & 13.9 & 63.3 \\
~~w/rand   & 42.0 & 24.7(4) & 49.5 & 8.7(1) & 33.5 & 10.6(4) & 37.9 & 14.7(1) & 40.7 & 14.7 & 63.1 \\ \bottomrule         
\end{tabular}

\caption{Correlation ($\tau$), inter-dimension dependence ($CG_\tau$), and number of dimensions whose inter-dimension dependence is statistically significant (in brackets) for DimCheck and its variant using random order to remove unrelated dimension. DimCheck's performance is stable with different prompts.}
\label{tab:dimcheck_order}
\end{table*}

From the above table, we find that changing the order can have some slight impact on each dataset, but the overall performance is stable, and DimCheck w/rand still shows consistent improvement. These findings suggest that DimCheck is not strongly sensitive to the order of dimensions when removing unrelated evidence.

Using Llama-3.3-70B on 4 NVIDIA RTX 6000 Ada GPUs, Analyze-Rate requires approximately 20 minutes to evaluate the SummEval dataset, while inference-time DimCheck requires approximately 120 minutes to edit all reasoning traces. To reduce this overhead, we train Llama-3.1-8B. Running on a single RTX 6000 Ada GPU, the trained variant completes the editing process in approximately 20 minutes, reducing the inference cost substantially while maintaining the benefits of DimCheck.

\subsection{COTs Editted by DimCheck}
\label{app:cot_edit}
\begin{figure*}
\centering
\includegraphics[width=\textwidth]{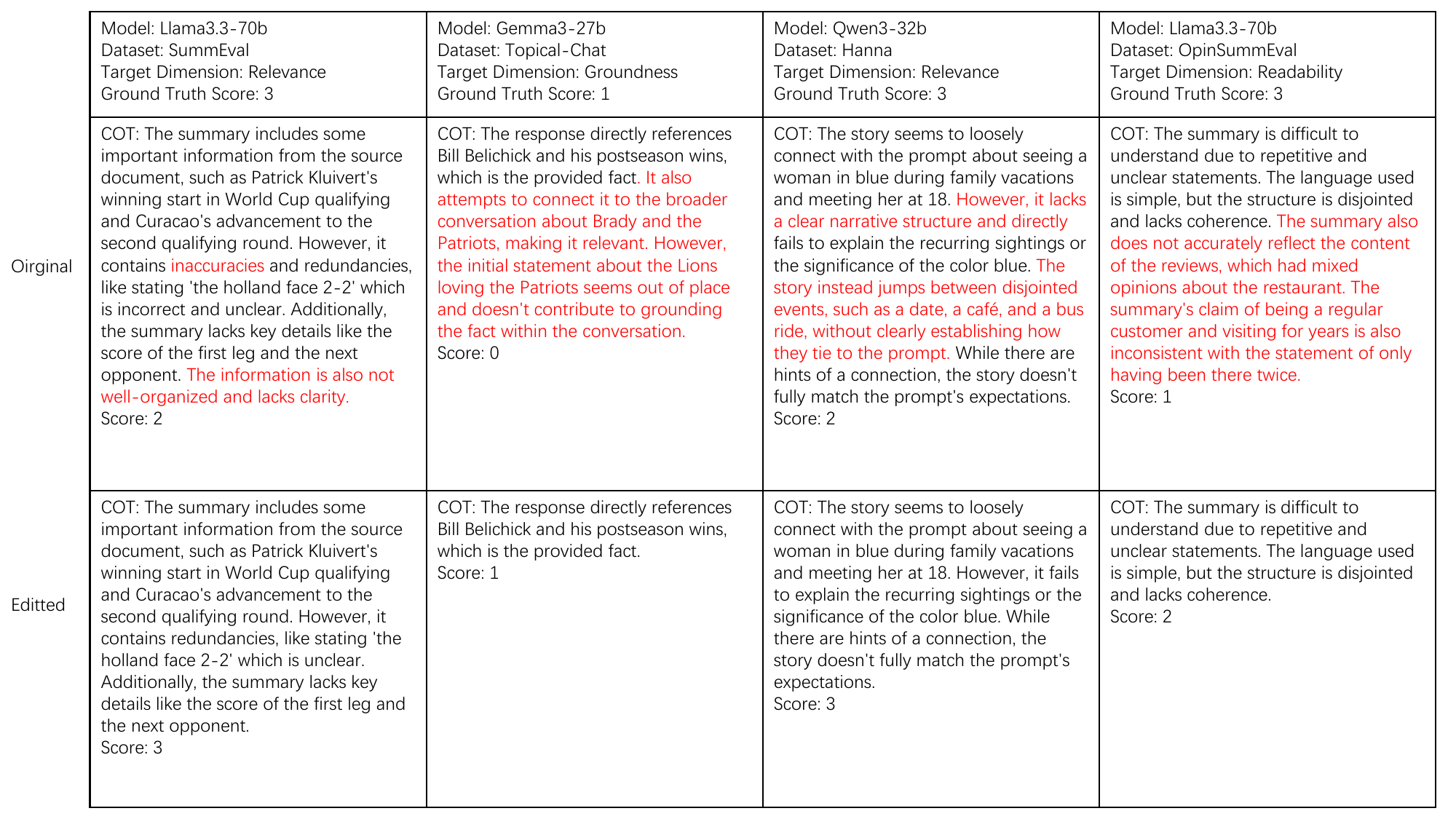}
\caption{Example COTs and scores edited by DimCheck. The evidence removed by DimCheck is in \textcolor{red}{red}. }
\label{fig:sample}
\end{figure*}  
\begin{table*}[]
\centering
\scriptsize
\begin{tabular}{lcccccccccc}
\toprule
             & \multicolumn{2}{c}{SummEval} & \multicolumn{2}{c}{Topical-Chat} & \multicolumn{2}{c}{Hanna} & \multicolumn{2}{c}{OpinSumm} & \multicolumn{2}{c}{Average} \\ 
             & COT          & Score         & COT            & Score           & COT         & Score       & COT          & Score         & COT          & Score        \\ \midrule
Llama3.3-70b & 46.3         & 8.8           & 53.4           & 3.7             & 60.7        & 3.3         & 63.9         & 15.9          & 56.1         & 7.9          \\
Gemma3-27b   & 84.1         & 14.1          & 66.0           & 3.6             & 89.6        & 5.7         & 96.3         & 16.3          & 84.0         & 9.9          \\
Qwen3-32b    & 54.4         & 11.4          & 65.9           & 2.6             & 78.3        & 6.9         & 59.8         & 12.1          & 64.6         & 8.2     \\ \bottomrule    
\end{tabular}
\caption{Proportion of COTs and scores edited by DimCheck.}
\label{tab：change_rate}
\end{table*}
We show example COTs edited by DimCheck and corresponding new scores in Fig. \ref{fig:sample}. We also report the proportion of COTs edited by DimCheck and proportion of changed score due to the edited COTs in Table. \ref{tab：change_rate}. As shown in the table, a large proportion of COTs is edited by DimCheck. As shown in Sec. \ref{sec:human_eval}, DimCheck can accurately identify COTs with unrelated evidence. The results show that a large proportion of COTs indeed contain unrelated evidence.

\begin{table*}[]
\centering
\scriptsize
\begin{tabular}{lcccccccccc}
\toprule
              & \multicolumn{2}{c}{SummEval} & \multicolumn{2}{c}{Topical Chat} & \multicolumn{2}{c}{Hanna} & \multicolumn{2}{c}{OpinSumm.} & \multicolumn{2}{c}{Average} \\
              & low           & high         & low             & high           & low         & high        & low           & high          & low          & high         \\ \midrule
              & \multicolumn{10}{c}{G-Eval}                                                                                                                               \\
Llama3.1-8b   & 9.5           & 23.1         & 19.8            & 10.0           & 8.8         & 7.9         & 9.9           & 6.2           & 12.0         & 11.8         \\
Llama3.3-70b  & 12.8          & 11.4         & 12.1            & 9.8            & 10.5        & 9.9         & 22.9          & 13.9          & 14.6         & 11.2         \\
Gemma3-12b    & 13.8          & 21.8         & 14.4            & 18.2           & 5.1         & 10.6        & 21.5          & 11.9          & 13.7         & 15.6         \\
Gemma3-27b    & 10.6          & 25.0         & 15.7            & 8.0            & 7.3         & 11.4        & 19.3          & 11.5          & 13.2         & 14.0         \\
Qwen3-14b     & 9.2           & 25.0         & 20.3            & 18.6           & 7.8         & 13.5        & 14.1          & 12.8          & 12.9         & 17.5         \\
Qwen3-32b     & 18.3          & 23.8         & 16.8            & 14.1           & 8.6         & 11.6        & 18.2          & 20.8          & 15.4         & 17.5         \\
Ministral-14b & 12.5          & 22.7         & 11.7            & 17.8           & 13.2        & 11.7        & 13.8          & 38.0          & 12.8         & 22.5         \\ \hdashline
              & \multicolumn{10}{c}{Analyze-Rate}                                                                                                                         \\
Llama3.1-8b   & 15.1          & 21.3         & 27.2            & 8.9            & 4.5         & 14.1        & 19.5          & 9.2           & 16.6         & 13.4         \\
Llama3.3-70b  & 10.6          & 32.2         & 16.0            & 15.5           & 9.7         & 14.9        & 14.1          & 23.9          & 12.6         & 21.6         \\
Gemma2-12b    & 13.4          & 26.2         & 19.4            & 9.5            & 6.5         & 12.9        & 15.4          & 24.2          & 13.7         & 18.2         \\
Gemma2-27b    & 10.5          & 30.1         & 14.3            & 3.8            & 9.4         & 10.7        & 18.8          & 26.4          & 13.2         & 17.7         \\
Qwen3-14b     & 13.2          & 28.2         & 13.0            & 18.8           & 10.3        & 16.8        & 21.0          & 21.0          & 14.3         & 21.2         \\
Qwen3-32b     & 10.8          & 24.0         & 28.1            & 8.7            & 9.8         & 15.7        & 11.7          & 25.9          & 15.1         & 18.6         \\
Ministral-14b & 20.0          & 24.5         & 10.2            & 15.3           & 6.2         & 13.2        & 17.0          & 23.8          & 13.4         & 19.2         \\
M-Prom.-14b   & 8.3           & 20.2         & 12.7            & 4.2            & 8.9         & 12.5        & 18.0          & 18.1          & 12.0         & 13.8         \\ \hdashline
              & \multicolumn{10}{c}{Debate}                                                                                                                               \\
Llama3.1-8b   & 16.5          & 17.7         & 12.8            & 12.8           & 4.2         & 7.2         & 21.6          & 11.2          & 13.8         & 12.2         \\
Llama3.3-70b  & 8.7           & 30.6         & 8.9             & 8.9            & 8.6         & 13.5        & 10.2          & 17.8          & 9.1          & 17.7         \\
Gemma3-12b    & 6.5           & 18.7         & 15.2            & 8.6            & 10.3        & 6.1         & 20.6          & 27.0          & 13.1         & 15.1         \\
Gemma3-27b    & 5.1           & 22.1         & 12.2            & 5.1            & 8.6         & 14.9        & 11.9          & 15.3          & 9.5          & 14.3         \\
Qwen3-14b     & 17.0          & 27.3         & 14.4            & 18.6           & 10.4        & 14.2        & 16.7          & 18.7          & 14.6         & 19.7         \\
Qwen3-32b     & 16.2          & 22.4         & 21.4            & 11.8           & 9.5         & 10.4        & 25.4          & 23.7          & 18.1         & 17.1         \\
Ministral-14b & 5.2           & 9.7          & 12.3            & 16.0           & 10.3        & 11.3        & 33.0          & 7.5           & 15.2         & 11.1     \\ \bottomrule   
\end{tabular}
\caption{Inter-dimension dependence on low-quality and high-quality groups. Inter-dimension dependence can vary across texts of different quality levels.}
\label{tab:corrgap_diff}
\end{table*}

\subsection{Implementation Detail of Human Evaluation}
\label{app:human_eval}

\begin{figure*}
\centering
\includegraphics[width=\textwidth]{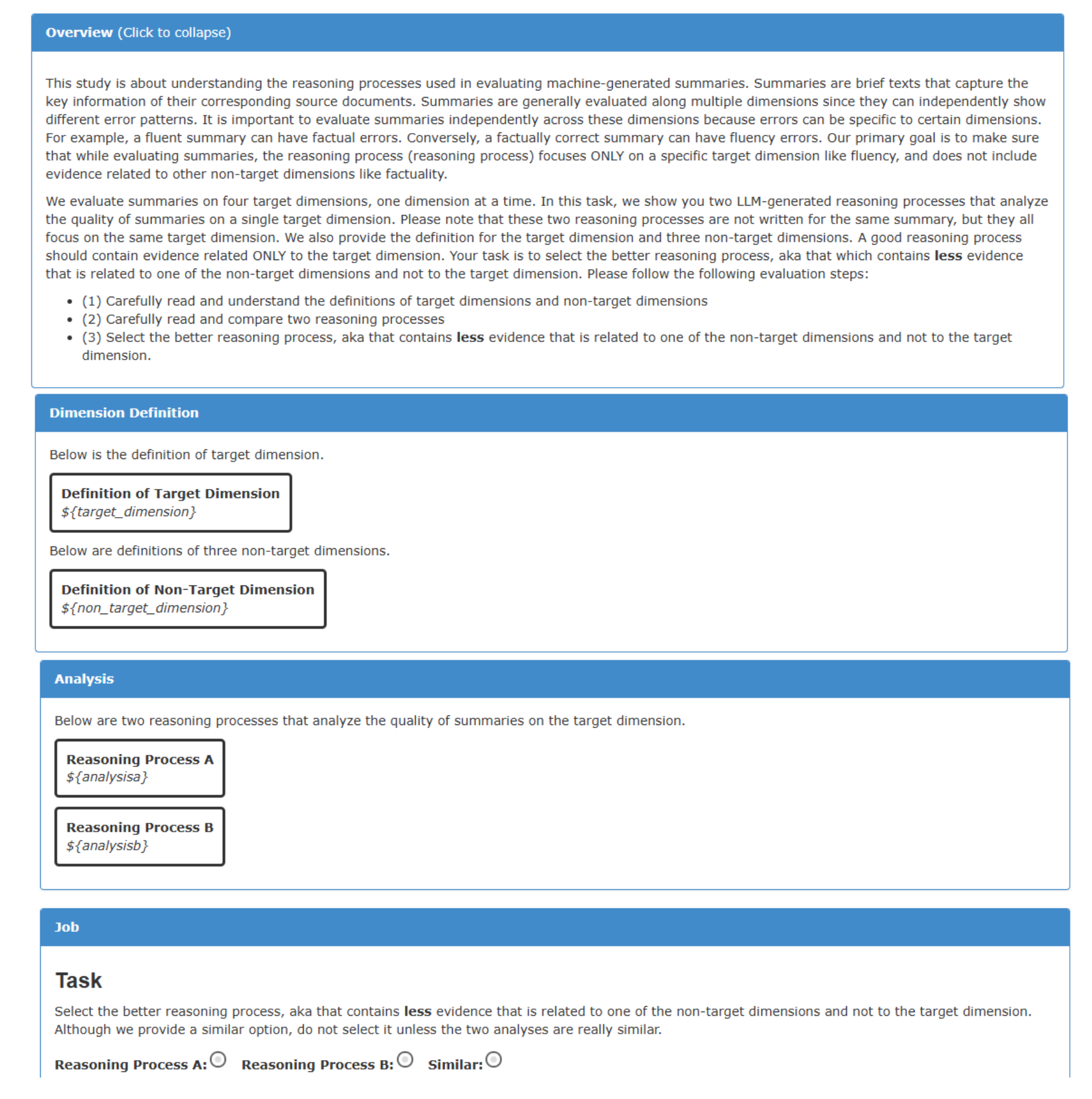}
\caption{Interface for Human Evaluation}
\label{fig:human}
\end{figure*}  

 Each human evaluation sample is annotated by three annotators. The annotators for human evaluation are recruited from Amazon Mechanical Turk. The annotators should be from English-speaking countries, have HIT Approval Rates greater than $98\%$ and number of HITS approved greater than 1000. The annotators should also be MTurk masters. The interface of human evaluation is shown in Fig. \ref{fig:human}.

\begin{table}[]
\centering
\scriptsize
\setlength{\tabcolsep}{1mm}
\begin{tabular}{lccccc}
\toprule
            & SummEval & Topical-Chat & Hanna & OpinSummEval & Average \\ \midrule
w/ control  & 0.00     & 0.00         & 0.00  & 0.31         & 0.08    \\
w/o control & 1.51     & 0.11         & 0.09  & 1.00         & 0.68   \\ \midrule
\end{tabular}
\caption{KL-divergence between the empirical distributions of low-variance group and high-variance group for CorrGap. Without controling, There can be big difference between the empirical distributions.}
\label{tab:kl_div}
\end{table}

\begin{figure*}[t]
\centering
\includegraphics[width=0.9\textwidth,keepaspectratio]{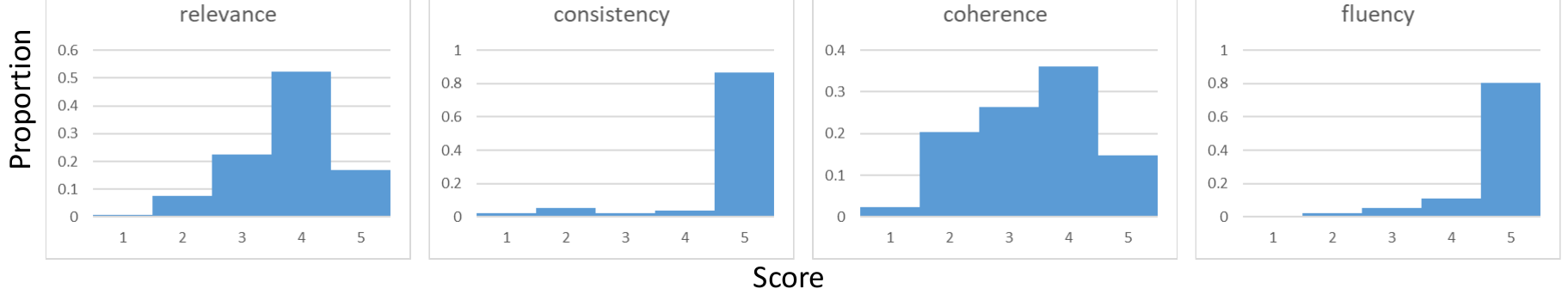}
\caption{Distribution of ground truth scores on different dimensions on the SummEval dataset. Ground truth scores across dimensions have very different distributions}
\label{fig:distribution}
\end{figure*}

\subsection{Implementation Details of LLM judges}
\label{app:llm_judge_detail}
The evaluation instructions of G-Eval and Analyze-Rate for SummEval and Topical-Chat datasets are from \citet{liu2023g}. The evaluation instruction for Hanna dataset is based on the `Eval-Prompt 3' which includes the rubric from \citet{chhun2024do}. The evaluation instruction for OpinSummEval datasets are from \citet{shen2023opinsummeval}. For debate, we use the `plain' version of the critic prompt from \citet{kim2024debate}, as we find it shows the best overall performance. 

\subsection{Implementation Details of Baselines}
\label{app:detail_baseline}
For Analyze-Rate+Joint, we show evaluation instructions for all dimensions and instruct LLMs to jointly predict scores for all dimension while ensuring the evaluation of each dimension is independent. Its prompt on the SummEval dataset is shown in Tab. \ref{tab:prompt_joint}. For Analyze-Rate+other, we additionally provide the definition of other dimensions and require the model's prediction should not be affected by these dimensions. Its prompt on the SummEval dataset for the relevance dimension is shown in Tab. \ref{tab:prompt_joint}. For self-consistency, we use the same prompt as Analyze-Rate but randomly sample COTs and scores for $5$ runs and use the majority of sampled scores as the final predicted score. For CheckEval, we only report the performance on the SummEval and Topical Chat datasets since it requires human-written seed questions for each dimension and the original paper only performs experiments on these two datasets. For self-reflection, we first run evaluation using the same prompt as Analyze-Rate and we then use a critic to decide whether the analysis and the score of the evaluation are affected by non-target dimensions and refine the score accordingly. The prompt used for the critic is shown in Tab. \ref{tab:prompt_reflection}. 

\begin{table*}[]
\centering
\scriptsize
\begin{tabular}{p{15cm}}
You will be given one summary written for a source document.                                                                                                                                                                                                                                                                                                 \\
Your task is to rate the summary on one metric.                                                                                                                                                                                                                                                                                                              \\
Please make sure you read and understand these instructions carefully. Please keep this document open while reviewing, and refer to it as needed.Evaluation Criteria:                                                                                                                                                                                        \\
Relevance (1-5) - selection of important content from the source. The summary should receive high score if it include only important information from the source document. Please penalize summaries which contained redundancies and excess information.                                                                                                    \\
Evaluation Steps:                                                                                                                                                                                                                                                                                                                                            \\
1. Read the summary and the source document carefully.                                                                                                                                                                                                                                                                                                       \\
2. Compare the summary to the source documents and identify the main points of the source documents.                                                                                                                                                                                                                                                         \\
3. Assess how well the summary covers the main points of the source documents, and how much irrelevant or redundant information it contains.                                                                                                                                                                                                                 \\
4. Assign a relevance score from 1 to 5.                                                                                                                                                                                                                                                                                                                     \\
Your evaluation should not be based on the following criteria:                                                                                                                                                                                                                                                                                               \\
Coherence (1-5) - the collective quality of all sentences. We align this dimension with the DUC quality question of structure and coherence whereby "the summary should be well-structured and well-organized. The summary should not just be a heap of related information, but should build from sentence to a coherent body of information about a topic. \\
Fluency (1-5): the quality of the summary in terms of grammar, spelling, punctuation, word choice, and sentence structure. You should not consider capitalization when making evaluation.                                                                                                                                                                    \\
Consistency (1-5) - the factual alignment between the summary and the summarized source. The summary should receive high score if it contains only statements that are entailed by the source document. Please penalize summaries that contained hallucinated facts.                                                                                        
\end{tabular}
\caption{Prompt for Analyze-Rate+Other on the SummEval dataset for the relevance dimension}
\label{tab:prompt_other}

\end{table*}

\begin{table*}[]
\scriptsize
\centering
\begin{tabular}{p{15cm}}
You will be given one summary written for a source document.                                                                                                                                                                                                                                                                                                 \\
Your task is to rate the summary on four metrics.                                                                                                                                                                                                                                                                                                            \\
Please make sure you read and understand these instructions carefully. Please keep this document open while reviewing, and refer to it as needed.                                                                                                                                                                                                            \\
You should evaluate each metric independently based on its corresponding criteria and evlauation steps.                                                                                                                                                                                                                                                      \\
Evaluation Criteria 1:                                                                                                                                                                                                                                                                                                                                       \\
Relevance (1-5) - selection of important content from the source. The summary should receive high score if it include only important information from the source document. Please penalize summaries which contained redundancies and excess information.                                                                                                    \\
Evaluation Steps 1:                                                                                                                                                                                                                                                                                                                                          \\
1. Read the summary and the source document carefully.                                                                                                                                                                                                                                                                                                       \\
2. Compare the summary to the source documents and identify the main points of the source documents.                                                                                                                                                                                                                                                         \\
3. Assess how well the summary covers the main points of the source documents, and how much irrelevant or redundant information it contains.                                                                                                                                                                                                                 \\
4. Generate an analysis of the summary and then assign a relevance score from 1 to 5.                                                                                                                                                                                                                                                                        \\
Evaluation Criteria 2:                                                                                                                                                                                                                                                                                                                                       \\
Consistency (1-5) - the factual alignment between the summary and the summarized source. The summary should receive high score if it contains only statements that are entailed by the source document. Please penalize summaries that contained hallucinated facts.                                                                                         \\
Evaluation Steps 2:                                                                                                                                                                                                                                                                                                                                          \\
1. Read the news article carefully and identify the main facts and details it presents.                                                                                                                                                                                                                                                                      \\
2. Read the summary and compare it to the article. Check if the summary contains any factual errors that are not supported by the article.                                                                                                                                                                                                                   \\
3. Generate an analysis of the summary and then assign a score for consistency from 1 to 5. based on the Evaluation Criteria.                                                                                                                                                                                                                                \\
Evaluation Criteria 3:                                                                                                                                                                                                                                                                                                                                       \\
Coherence (1-5) - the collective quality of all sentences. We align this dimension with the DUC quality question of structure and coherence whereby "the summary should be well-structured and well-organized. The summary should not just be a heap of related information, but should build from sentence to a coherent body of information about a topic. \\
Evaluation Steps 3:                                                                                                                                                                                                                                                                                                                                          \\
1. Read the news article carefully and identify the main topic and key points.                                                                                                                                                                                                                                                                               \\
2. Read the summary and compare it to the news article. Check if the summary presents the key points in a clear and logical order.                                                                                                                                                                                                                           \\
3. Generate an analyisis of the summary and then assign a score for coherence on a scale of 1 to 5, where 1 is the lowest and 5 is the highest based on the Evaluation Criteria.                                                                                                                                                                             \\
Evaluation Criteria 4:                                                                                                                                                                                                                                                                                                                                       \\
Fluency (1-5): the quality of the summary in terms of grammar, spelling, punctuation, word choice, and sentence structure. You should not consider capitalization when making evaluation.                                                                                                                                                                    \\
Evaluation Steps 4:                                                                                                                                                                                                                                                                                                                                          \\
1. Read the summary carefully and identify errors that make it hard to understand or sound unnatural.                                                                                                                                                                                                                                                        \\
2. Generate an analysis of the summary and then assign a score for fluency on a scale of 1 to 5, where 1 is the lowest and 5 is the highest based on the Evaluation Criteria.                                                                                                                                                                         
\end{tabular}
\caption{Prompt for Analyze-Rate+Joint on the SummEval dataset}
\label{tab:prompt_joint}
\end{table*}

\begin{table*}[]
\scriptsize
\centering
\begin{tabular}{{p{15cm}}}
You will be given a reasoning process generated by an LLM judge that is instructed to evaluate a summary on a target dimension. You will also be given evaluation instructions on the target dimension and three non-target dimensions.                                                                                                                                                                                                                                                                                                                                                                                                                                                                                                                                                                               \\
Your task is to first check the reasoning process to find out whether it contains evidence that is closely related to any of the three non-target dimensions but not related to the target dimension. For this step, you should not consider any dimensions other than the given three non-target dimensions.                                                                                                                                                                                                                                                                                                                                                                                                                                                                                                         \\
If the reasoning process contains such evidence, you should remove all such evidence from the reasoning process. You should only make necessary modifications without changing other parts of the reasoning process. You should not add content that is not already presented in the original reasoning process. You should then return the modified reasoning process without extra comments. If you do not find such evidence or all contents are moved, return 'NO CHANGE'.                                                                                                                                                                                                                                                                                                                                        \\
Reasoning Process:                                                                                                                                                                                                                                                                                                                                                                                                                                                                                                                                                                                                                                                                                                                                                                                                    \\
\{reasoning\_process\} \\
Evaluation instruction for the target dimension:                                                                                                                                                                                                                                                                                                                                                                                                                                                                                                                                                                                                                                                                                                                                                                      \\
You will be given one summary written for a source document.                                                                                                                                                                                                                                                                                                                                                                                                                                                                                                                                                                                                                                                                                                                                                          \\
Your task is to rate the summary on one metric.                                                                                                                                                                                                                                                                                                                                                                                                                                                                                                                                                                                                                                                                                                                                                                       \\
Please make sure you read and understand these instructions carefully. Please keep this document open while reviewing, and refer to it as needed.Evaluation Criteria:                                                                                                                                                                                                                                                                                                                                                                                                                                                                                                                                                                                                                                                 \\
Relevance (1-5) - selection of important content from the source. The summary should receive high score if it include only important information from the source document. Please penalize summaries which contained redundancies and excess information.                                                                                                                                                                                                                                                                                                                                                                                                                                                                                                                                                             \\
Evaluation Steps:                                                                                                                                                                                                                                                                                                                                                                                                                                                                                                                                                                                                                                                                                                                                                                                                     \\
1. Read the summary and the source document carefully.                                                                                                                                                                                                                                                                                                                                                                                                                                                                                                                                                                                                                                                                                                                                                                \\
2. Compare the summary to the source documents and identify the main points of the source documents.                                                                                                                                                                                                                                                                                                                                                                                                                                                                                                                                                                                                                                                                                                                  \\
3. Assess how well the summary covers the main points of the source documents, and how much irrelevant or redundant information it contains.                                                                                                                                                                                                                                                                                                                                                                                                                                                                                                                                                                                                                                                                          \\
4. Generate an analysis of the summary and then assign a relevance score from 1 to 5.                                                                                                                                                                                                                                                                                                                                                                                                                                                                                                                                                                                                                                                                                                                                 \\
Evaluation instruction for non-target dimension 1:                                                                                                                                                                                                                                                                                                                                                                                                                                                                                                                                                                                                                                                                                                                                                                    \\
You will be given one summary written for a source document.                                                                                                                                                                                                                                                                                                                                                                                                                                                                                                                                                                                                                                                                                                                                                          \\
Your task is to rate the summary on one metric.                                                                                                                                                                                                                                                                                                                                                                                                                                                                                                                                                                                                                                                                                                                                                                       \\
Please make sure you read and understand these instructions carefully. Please keep this document open while reviewing, and refer to it as needed.Evaluation Criteria:                                                                                                                                                                                                                                                                                                                                                                                                                                                                                                                                                                                                                                                 \\
Consistency (1-5) - the factual alignment between the summary and the summarized source. The summary should receive high score if it contains only statements that are entailed by the source document. Please penalize summaries that contained hallucinated facts.                                                                                                                                                                                                                                                                                                                                                                                                                                                                                                                                                  \\
Evaluation Steps:                                                                                                                                                                                                                                                                                                                                                                                                                                                                                                                                                                                                                                                                                                                                                                                                     \\
1. Read the news article carefully and identify the main facts and details it presents.                                                                                                                                                                                                                                                                                                                                                                                                                                                                                                                                                                                                                                                                                                                               \\
2. Read the summary and compare it to the article. Check if the summary contains any factual errors that are not supported by the article.                                                                                                                                                                                                                                                                                                                                                                                                                                                                                                                                                                                                                                                                            \\
3. Generate an analysis of the summary and then assign a score for consistency from 1 to 5. based on the Evaluation Criteria.                                                                                                                                                                                                                                                                                                                                                                                                                                                                                                                                                                                                                                                                                         \\
Evaluation instruction for non-target dimension 2:                                                                                                                                                                                                                                                                                                                                                                                                                                                                                                                                                                                                                                                                                                                                                                    \\
You will be given one summary written for a source document.                                                                                                                                                                                                                                                                                                                                                                                                                                                                                                                                                                                                                                                                                                                                                          \\
Your task is to rate the summary on one metric.                                                                                                                                                                                                                                                                                                                                                                                                                                                                                                                                                                                                                                                                                                                                                                       \\
Please make sure you read and understand these instructions carefully. Please keep this document open while reviewing, and refer to it as needed.Evaluation Criteria:                                                                                                                                                                                                                                                                                                                                                                                                                                                                                                                                                                                                                                                 \\
Coherence (1-5) - the collective quality of all sentences. We align this dimension with the DUC quality question of structure and coherence whereby "the summary should be well-structured and well-organized. The summary should not just be a heap of related information, but should build from sentence to a coherent body of information about a topic.                                                                                                                                                                                                                                                                                                                                                                                                                                                          \\
Evaluation Steps:                                                                                                                                                                                                                                                                                                                                                                                                                                                                                                                                                                                                                                                                                                                                                                                                     \\
1. Read the news article carefully and identify the main topic and key points.                                                                                                                                                                                                                                                                                                                                                                                                                                                                                                                                                                                                                                                                                                                                        \\
2. Read the summary and compare it to the news article. Check if the summary presents the key points in a clear and logical order.                                                                                                                                                                                                                                                                                                                                                                                                                                                                                                                                                                                                                                                                                    \\
3. Generate an analyisis of the summary and then assign a score for coherence on a scale of 1 to 5, where 1 is the lowest and 5 is the highest based on the Evaluation Criteria.                                                                                                                                                                                                                                                                                                                                                                                                                                                                                                                                                                                                                                      \\
Evaluation instruction for non-target dimension 3:                                                                                                                                                                                                                                                                                                                                                                                                                                                                                                                                                                                                                                                                                                                                                                    \\
You will be given one summary written for a source document.                                                                                                                                                                                                                                                                                                                                                                                                                                                                                                                                                                                                                                                                                                                                                          \\
Your task is to rate the summary on one metric.                                                                                                                                                                                                                                                                                                                                                                                                                                                                                                                                                                                                                                                                                                                                                                       \\
Please make sure you read and understand these instructions carefully. Please keep this document open while reviewing, and refer to it as needed.Evaluation Criteria:                                                                                                                                                                                                                                                                                                                                                                                                                                                                                                                                                                                                                                                 \\
Fluency (1-5): the quality of the summary in terms of grammar, spelling, punctuation, word choice, and sentence structure. You should not consider capitalization when making evaluation.                                                                                                                                                                                                                                                                                                                                                                                                                                                                                                                                                                                                                             \\
Evaluation Steps:                                                                                                                                                                                                                                                                                                                                                                                                                                                                                                                                                                                                                                                                                                                                                                                                     \\
1. Read the summary carefully and identify errors that make it hard to understand or sound unnatural.                                                                                                                                                                                                                                                                                                                                                                                                                                                                                                                                                                                                                                                                                                                 \\
2. Generate an analysis of the summary and then assign a score for fluency on a scale of 1 to 5, where 1 is the lowest and 5 is the highest based on the Evaluation Criteria.                                    
\end{tabular}
\caption{Prompt for SFT on the SummEval dataset}
\label{tab:prompt_sft}
\end{table*}

\begin{table*}[]
\scriptsize
\centering
\begin{tabular}{lcccccc}
\toprule
             & Relevance        & Consistency    & Coherence             & Fluency     &            &            \\
SummEval     & 22.4             & 20.5           & 31.8                  & 34.2        &            &            \\ \hdashline
             & Engaging         & Natural        & Coherence             & Groundness  &            &            \\
Topical Chat & 1.5              & 14.4           & 13.9                  & 12.0        &            &            \\ \hdashline
             & Relevance        & Coherence      & Empathy               & Surprise    & Engagement & Complexity \\
Hanna        & 5.2              & 5.4            & 5.6                   & 12.0        & 17.9       & 15.1       \\ \hdashline
             & Aspect Relevance & Self Coherence & Sentiment Consistency & Readability &            &            \\ 
OpinSummEval & 32.6             & 12.8           & 12.2                  & 13.8        &            &         \\
\bottomrule
\end{tabular}
\caption{Average CorrGap when using Analyze-rate with Llama-3.3-70B-Instruct, Gemma-3-27B-it, and Qwen3-32B for each evaluation dimension. Different dimensions show very different CorrGap values.}
\label{tab:dimension_analysis}
\end{table*}

\subsection{Prompt Stability of DimCheck}
\label{app:dimcheck_stable}
\begin{table*}[]
\centering
\scriptsize
\setlength{\tabcolsep}{1mm}
\begin{tabular}{lccccccccccc}
\toprule
                   & \multicolumn{2}{c}{SummEval[4]} & \multicolumn{2}{c}{Topical Chat[4]} & \multicolumn{2}{c}{Hanna[6]} & \multicolumn{2}{c}{OpinSumm.[4]} & \multicolumn{2}{c}{Avg.}                         \\
                  & $\tau\uparrow$             &  CG$\downarrow$         & $\tau\uparrow$               & CG$\downarrow$             & $\tau\uparrow$           & CG$\downarrow$          & $\tau\uparrow$               & CG$\downarrow$             & $\tau\uparrow$              & CG$\downarrow$             & overall$\uparrow$        \\ \midrule
                   & \multicolumn{11}{c}{Llama3.3-70b}                                                                                                                                                 \\ 
Analyze-Rate       & \textbf{46.1} & 30.0(4)          & 53.3          & 13.1(3)          & 34.6          & 11.3(3)          & 32.0          & 18.9(3)          & 41.5          & 18.3          & 61.6          \\
Analyze-Rate+Other & 45.1          & 30.5(4)          & 52.0          & 13.8(3)          & 35.9 & 12.6(3)          & \textbf{34.9} & 17.4(2)          & 42.0          & 18.6          & 61.7          \\
Analyze-Rate+Joint & 41.4 & 28.6(4) & 53.5 & 13.9(2) & \textbf{37.2} & 13.7(6) & 33.4 & 19.5(2) & 41.4 & 18.9 & 61.2 \\ 
Analyze-Rate+DimCheck     & 45.9          & \textbf{27.9}(4) & \textbf{54.1} & \textbf{12.5}(3) & 35.0          & \textbf{11.1}(3) & 33.2          & \textbf{14.1}(1) & \textbf{42.0} & \textbf{16.4} & \textbf{62.8}             \\
~~+DimCheck w/var.  & 45.2 & 28.4(4) & 54.3 & 12.3(3) & 35.4 & 10.4(3) & 33.5 & 16.2(1) & 42.1 & 16.8 & 62.6 \\
 \hdashline
                   & \multicolumn{11}{c}{Gemma3-27b}                                                                                                                                                                                                                                           \\ 
Analyze-Rate       & 44.1          & 26.3(4)          & 51.1          & 10.7(2)         & 28.7          & 11.4(3)         & 37.6          & 16.1(2)          & 40.4          & 16.1          & 62.1          \\
Analyze-Rate+Other & \textbf{45.6} & 27.0(4)          & 49.9          & 12.6(2)         & 29.8 & \textbf{8.5}(2) & \textbf{38.0} & 17.2(2)          & \textbf{40.8} & 16.3          & 62.3          \\
Analyze-Rate+Joint & 43.0 & 23.6(4) & 47.6 & 12.8(1) & \textbf{32.3} & 10.5(4) & 35.7 & \textbf{14.1}(1) & 39.6 & 15.2 & 62.2 \\
Analyze-Rate+DimCheck     & 42.9          & \textbf{22.3}(4) & \textbf{51.3} & \textbf{9.5}(2) & 29.2          & 10.1(3)         & 38.0          & 15.0(1) & 40.3          & \textbf{14.2} & \textbf{63.1}             \\
~~+DimCheck w/var. & 43.0 & 22.6(4) & 51.4 & 9.1(2) & 29.1 & 10.3(3) & 37.0 & 14.7(1) & 40.1 & 14.1 & 63.0 \\
 \hdashline

                   & \multicolumn{11}{c}{Qwen3-32b}                                                                                                                                                                                                                                                   \\
Analyze-Rate       & 41.8          & 25.3(4)          & 49.1          & 10.1(1)         & 34.3          & 11.7(4)         & \textbf{37.6} & 18.5(3)          & \textbf{40.7} & 16.4          & 62.1          \\
Analyze-Rate+Other & 41.0          & 27.6(4)          & 48.5          & 10.5(2)         & \textbf{35.6} & \textbf{9.4}(3) & 26.0          & 17.0(2)          & 37.8          & 16.1          & 60.8          \\
Analyze-Rate+Joint & \textbf{42.2} & 29.1(4)  & \textbf{52.0} & 17.2(4) & 34.1 & 14.2(5) & 35.4 & 21.7(2) & 40.9 & 20.5 & 60.2 \\
Analyze-Rate+DimCheck     & 42.0 & \textbf{23.9}(4) & 49.9 & \textbf{8.5}(1) & 33.7          & 10.9(4)         & 35.9          & \textbf{12.2}(1) & 40.4          & \textbf{13.9} & \textbf{63.3} \\ 
~~+DimCheck w/var. & 41.8 & 25.0(4) & 49.5 & 8.8(1) & 34.0 & 10.9(4) & 37.2 & 12.6(1) & 40.6 & 14.3 & 63.1 \\\bottomrule         
\end{tabular}

\caption{Correlation ($\tau$), inter-dimension dependence ($CG_\tau$), and number of dimensions whose inter-dimension dependence is statistically significant (in brackets) for DimCheck and its variant using different prompts. The best-performing method is \textbf{bolded}. DimCheck's performance is stable with different prompts.}
\label{tab:dimcheck_stable}
\end{table*}
\begin{table*}[]
\centering
\scriptsize
\begin{tabular}{p{15cm}}
You will be given a reasoning process generated by an LLM judge that is instructed to evaluate a \{output\_type\} on a target dimension. You will also be given evaluation instructions on the target dimension and a non-target dimension.                                                                                                                                                                                                    \\
Your task is to first check the reasoning process to identify evidence that is closely related to the non-target dimension but not related to the target dimension. For this step, you should not consider any dimensions other than the given target dimension and non-target dimension.                                                                                                                               \\
If you do not find such evidence, return 'NO CHANGE'. \\  
If the reasoning process contains such evidence, you should remove all such evidence from the reasoning process. You should only make necessary modifications without changing other parts of the reasoning process. You should not add content that is not already presented in the original reasoning process. You should then return the modified reasoning process without extra comments. If all contents are removed, return 'EMPTY'. \\

Reasoning Process:                                                                                                                                                                                                                                                                                                                                                                                                                          \\
\{reasoning\_process\}                                                                                                                                                                                                                                                                                                                                                                                                                         \\
Evaluation instruction for the target dimension:                                                                                                                                                                                                                                                                                                                                                                                            \\
\{evaluation\_instruction\_target\}                                                                                                                                                                                                                                                                                                                                                                                                             \\
Evaluation instruction for the non-target dimension:                                                                                                                                                                                                                                                                                                                                                                                        \\
\{evaluation\_instruction\_non\_target\}                                                                                                                                                                                                                                                                                                                                                                                                         \\
You should first think step-by-step and then return in the following format:                                                                                                                                                                                                                                                                                                                                                                \\
THINK: ...                                                                                                                                                                                                                                                                                                                                                                                                                                  \\
ANSWER: modified reasoning process or 'EMPTY' or 'NO CHANGE'                                                                                                                                                                                                                                                                                                                                                                  
\end{tabular}
\caption{A variant prompt for DimCheck}
\label{tab:prompt_var}
\end{table*}

To evaluate the prompt stability of DimCheck, we report its performance on a variant prompt we previously tried (DimCheck w/var.). The variant prompt is in Tab. \ref{tab:prompt_var}. Compared with the prompt used by DimCheck shown in Tab. \ref{tab:prompt_dimcheck}, the variant prompt has different phrases and reverse the order of some instructions. We report the results in Tab. \ref{tab:dimcheck_stable}.

From the table, we observe that the performance of DimCheck w/var. is very close to DimCheck. Besides, the performance difference between DimCheck w/var. and the best-performing other method is statistically significant ($p<0.05$) using bootstrapping test. The results show that DimCheck's performance is stable with different prompts.

\subsection{Inter-Dimension Dependence for Different Scores}
\label{app:corr_diff}
We analyze whether inter-dimension dependence varies across texts of different quality levels. Specifically, for each dimension, we split all texts into low-quality and high-quality groups based on whether their ground-truth scores are low or high. The low-quality group contains texts whose ground-truth scores are $\leq2$  for the Topical Chat dataset since its scoring range is 1 to 3 and $\leq3$ for other datasets since their scoring range is 1 to 5. 
The high-quality group contains the remaining text. For this analysis, we do not consider the groundingness dimension of the Topical Chat dataset due to its overly coarse scoring range (0–1). 
For each group, we report the average inter-dimension dependence measured by CorrGap across LLMs for each evaluation framework in Tab. \ref{tab:corrgap_diff}.

From the table, we find that inter-dimension dependence is more severe on the high-quality groups on the SummEval and Hanna datasets while more severe on the low-quality groups on the Topical Chat dataset. The results show that inter-dimension dependence can vary across texts of different quality levels. 

\begin{table*}[]
\centering
\scriptsize
\begin{tabular}{lccccccccccc}
\toprule
                   & \multicolumn{2}{c}{SummEval} & \multicolumn{2}{c}{Topical Chat} & \multicolumn{2}{c}{Hanna} & \multicolumn{2}{c}{OpinSumm.} & \multicolumn{2}{c}{Avg.}                         \\
 & $\tau\uparrow$             &  CG$\downarrow$         & $\tau\uparrow$               & CG$\downarrow$             & $\tau\uparrow$           & CG$\downarrow$          & $\tau\uparrow$               & CG$\downarrow$             & $\tau\uparrow$              & CG$\downarrow$             & overall$\uparrow$        \\ \midrule
                   & \multicolumn{11}{c}{Llama3.3-70b}                                                                                                                                                 \\ 
DimCheck     & 45.9          & \textbf{27.9}(4) & \textbf{54.1} & \textbf{12.5}(3) & 35.0          & \textbf{11.1}(3) & 33.2          & \textbf{14.1}(1) & \textbf{42.0} & \textbf{16.4} & \textbf{62.8}             \\  
DimCheck w/1 step & 45.2 & 28.5(4) & 53.8 & 11.9(3) & 34.9 & 10.2(3) & 34.1 & 14.7(1) & 42.0 & 16.3 & 62.8 \\
DimCheck   w/o other & 46.0 & 29.3(4) & 53.4 & 12.3(3) & 35.0 & 10.7(3) & 33.4 & 15.6(1) & 42.0 & 16.9 & 62.5 \\ 
DimCheck   w/o think & 44.2 & 28.1(4) & 52.8 & 12.9(2) & 34.9 & 10.0(3) & 35.1 & 15.5(1) & 41.7 & 16.6 & 62.5 \\ \hdashline
                   & \multicolumn{11}{c}{Gemma3-27b}                                                                                                                                                                                                                                           \\ 

DimCheck     & 42.9          & \textbf{22.3}(4) & \textbf{51.3} & \textbf{9.5}(2) & 29.2          & 10.1(3)         & 38.0          & 15.0(1) & 40.3          & \textbf{14.2} & \textbf{63.1}             \\ 
DimCheck   w/1 step & 43.0 & 24.1(4) & 51.7 & 9.7(1) & 29.5 & 10.2(3) & 37.4 & 18.3(1) & 40.4 & 15.6 & 62.4 \\
DimCheck   w/o other & 43.9 & 24.5(4) & 51.3 & 9.8(1) & 28.8 & 11.3(3) & 37.7 & 17.9(1) & 40.4 & 15.9 & 62.3 \\ 
DimCheck   w/o think & 44.4 & 25.6(4) & 51.4 & 9.7(2) & 28.3 & 11.2(3) & 36.1 & 15.4(1) & 40.1 & 15.5 & 62.3 \\ \hdashline

                   & \multicolumn{11}{c}{Qwen3-32b}                                                                                                                                                                                                                                                   \\
DimCheck     & 42.0 & \textbf{23.9}(4) & 49.9 & \textbf{8.5}(1) & 33.7          & 10.9(4)         & 35.9          & \textbf{12.2}(1) & 40.4          & \textbf{13.9} & \textbf{63.3} \\
DimCheck w/1 step & 41.7 & 24.4(4) & 50.1 & 9.5(1) & 33.7 & 11.5(4) & 35.7 & 16.9(1) & 40.3 & 15.6 & 62.4 \\ 
DimCheck   w/o other & 40.9 & 25.9(4) & 49.3 & 9.7(1) & 34.4 & 10.9(3) & 38.6 & 16.5(1) & 40.8 & 15.8 & 62.5 \\ 
DimCheck   w/o think & 42.0 & 26.4(4) & 49.4 & 8.9(1) & 34.4 & 11.2(3) & 37.2 & 13.7(1) & 40.8 & 15.1 & 62.8 \\ \bottomrule         
\end{tabular}
\caption{Correlation ($\tau$), inter-dimension dependence ($CG_\tau$), and number of dimensions whose inter-dimension dependence is statistically significant (in brackets) for DimCheck and its ablated variants. The best-performing method is \textbf{bolded}. DimCheck outperforms all its ablated variants. }
\label{tab:ablation_eval}
\end{table*}

\begin{table*}[]
\centering
\scriptsize
\begin{tabular}{p{15cm}}
You will be given one summary written for a source document.                                                                                                                                                                                                                                                                                                 \\
You will also be given your previous evaluations.                                                                                                                                                                                                                                                                                                            \\
Your task is to decide whether your previous evaluation is affected by other dimensions and refine your score if it is affected by other dimensions.                                                                                                                                                                                                         \\
Evaluation Criteria:                                                                                                                                                                                                                                                                                                                                         \\
Relevance (1-5) - selection of important content from the source. The summary should receive high score if it include only important information from the source document. Please penalize summaries which contained redundancies and excess information.                                                                                                    \\
Evaluation Steps:                                                                                                                                                                                                                                                                                                                                            \\
1. Read the summary and the source document carefully.                                                                                                                                                                                                                                                                                                       \\
2. Compare the summary to the source documents and identify the main points of the source documents.                                                                                                                                                                                                                                                         \\
3. Assess how well the summary covers the main points of the source documents, and how much irrelevant or redundant information it contains.                                                                                                                                                                                                                 \\
4. Assign a relevance score from 1 to 5.                                                                                                                                                                                                                                                                                                                     \\
Other Dimensions:                                                                                                                                                                                                                                                                                                                                            \\
Coherence (1-5) - the collective quality of all sentences. We align this dimension with the DUC quality question of structure and coherence whereby "the summary should be well-structured and well-organized. The summary should not just be a heap of related information, but should build from sentence to a coherent body of information about a topic. \\
Fluency (1-5): the quality of the summary in terms of grammar, spelling, punctuation, word choice, and sentence structure. You should not consider capitalization when making evaluation.                                                                                                                                                                    \\
Consistency (1-5) - the factual alignment between the summary and the summarized source. The summary should receive high score if it contains only statements that are entailed by the source document. Please penalize summaries that contained hallucinated facts.                                                                                        
\end{tabular}
\caption{Prompt for Self-Reflection on the SummEval dataset for the relevance dimensions. }
\label{tab:prompt_reflection}
\end{table*}

\subsection{Implementation Details of Training Smaller LLMs for Efficient COT Editing}
\label{app:sft_detail}
We train models using the SFTTrainer from the TRL package with a learning rate of 1e-4, batch size of 16, and one training epoch. We additionally apply LoRA \cite{hu2022lora} fine-tuning with rank 32 and alpha 32, while using the default values of SFTTrainer for other hyperparameters. The SFT instruction follows the DimCheck prompt format, but additionally includes definitions of all non-target dimensions and requires the model to directly generate the revised reasoning process without COT as in Tab. \ref{tab:prompt_sft}.

As shown in Tab. \ref{tab:train_eval}, the trained version of DimCheck with smaller LLMs shows performance close to the inference version using larger LLMs. Furthermore, the trained approach significantly reduces inference cost. For example, the trained Llama-3.1-8B-Instruct model running on a single GPU reduces inference time by approximately 85\% compared with the inference-based version using Llama-3.3-70B-Instruct across four GPUs.

\end{document}